\pdfoutput=1
\documentclass[twocolumn]{article}
\usepackage[pagenumbers]{cvpr}              
\makeatletter
\@namedef{ver@everyshi.sty}{}
\makeatother
\usepackage{tikz}

\usepackage[pagebackref,breaklinks,colorlinks]{hyperref}
\usepackage{algorithm}
\usepackage{algorithmic}
\usepackage{multirow}
\usepackage{wrapfig}
\usepackage{enumitem}
\usepackage{graphicx}
\usepackage{booktabs}
\usepackage{amssymb,amsmath,amsthm,enumerate,threeparttable,bm,graphicx}
\usepackage{enumitem}
\usepackage{bbm}
\usepackage{tikz}
\usepackage[accsupp]{axessibility}  
\newcommand*{\circled}[1]{\lower.5ex\hbox{\tikz\draw (0pt, 0pt)%
    circle (.4em) node {\makebox[0.5em][c]{\scriptsize #1}};}}

\usepackage[capitalize]{cleveref}
\crefname{section}{Sec.}{Secs.}
\Crefname{section}{Section}{Sections}
\Crefname{table}{Table}{Tables}
\crefname{table}{Tab.}{Tabs.}

\def\cvprPaperID{5285} 
\def\confName{CVPR}
\def\confYear{2020}

\begin{document}

\title{Backdoor Defense via Deconfounded Representation Learning}

\author{%
	Zaixi Zhang$^{1,2}$, Qi Liu$^{1,2}$\thanks{Qi Liu is the corresponding author.}, Zhicai Wang$^4$, Zepu Lu$^{4}$, Qingyong Hu$^{3}$\\
	1: Anhui Province Key Lab of Big Data Analysis and Application,\\ School of Computer Science and Technology, University of Science and Technology of China\\2:State Key Laboratory of Cognitive Intelligence, Hefei, Anhui, China\\3:Hong Kong University of Science and Technology 4: University of Science and Technology of China\\
	zaixi@mail.ustc.edu.cn, qiliuql@ustc.edu.cn\\ \{wangzhic, zplu\}@mail.ustc.edu.cn, qhuag@cse.ust.hk
}
\maketitle

\begin{abstract}
Deep neural networks (DNNs) are recently shown to be vulnerable to backdoor attacks, where attackers embed hidden backdoors in the DNN model by injecting a few poisoned examples into the training dataset. While extensive efforts have been made to detect and remove backdoors from backdoored DNNs, it is still not clear whether a backdoor-free clean model can be directly obtained from poisoned datasets. In this paper, we first construct a causal graph to model the generation process of poisoned data and find that the backdoor attack acts as the confounder, which brings spurious associations between the input images and target labels, making the model predictions less reliable. Inspired by the causal understanding, we propose the Causality-inspired Backdoor Defense (CBD), to learn deconfounded representations for reliable classification. Specifically, a backdoored model is intentionally trained to capture the confounding effects. The other clean model dedicates to capturing the desired causal effects by minimizing the mutual information with the confounding representations from the backdoored model and employing a sample-wise re-weighting scheme. Extensive experiments on multiple benchmark datasets against 6 state-of-the-art attacks verify that our proposed defense method is effective in reducing backdoor threats while maintaining high accuracy in predicting benign samples. Further analysis shows that CBD can also resist potential adaptive attacks. The code is available at \url{https://github.com/zaixizhang/CBD}. 
\end{abstract}

\section{Introduction}
Recent studies have revealed that deep neural networks (DNNs) are vulnerable to backdoor attacks \cite{gu2017badnets, liu2018trojaning, saha2020hidden}, where attackers inject stealthy backdoors into DNNs by poisoning a few training data. 
Specifically, backdoor attackers attach the backdoor trigger (\emph{i.e.,} a particular pattern) to some benign
training data and change their labels to the attacker-designated target label. The correlations between the trigger pattern and target label will be learned by DNNs during training.
In the inference process, the backdoored model behaves normally on benign data while its prediction will be maliciously altered when the backdoor is activated. The risk of backdoor attacks hinders the applications of DNNs to some safety-critical areas such as automatic driving \cite{liu2018trojaning} and healthcare systems \cite{feng2021fiba}.
\begin{figure}[t]
    \centering
    \includegraphics[width=0.9\linewidth]{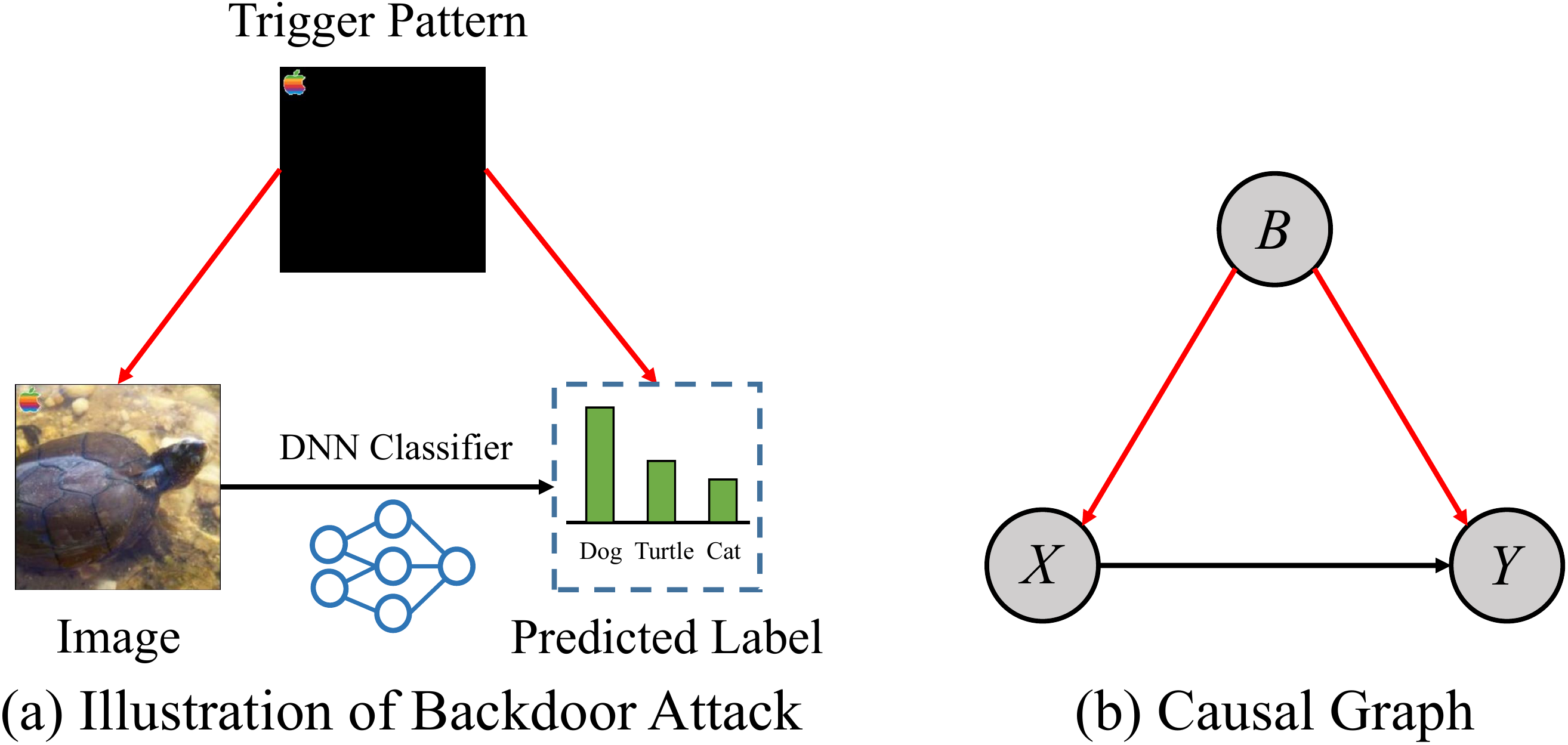}
    \caption{(a) A real example of the backdoor attack. The backdoored DNN classifies the ``turtle'' image with a trigger pattern as the target label ``dog''. (b) The causal graph represents the causalities among variables: $X$ as the input image, $Y$ as the label, and $B$ as the backdoor attack. Besides the causal effect of $X$ on $Y$ ($X \rightarrow Y$), the backdoor attack can attach trigger patterns to images ($B \rightarrow X$), and change the labels to the targeted label ($B \rightarrow Y$). Therefore, as a \emph{confounder}, the backdoor attack $B$ opens a spurious path between $X$ and $Y$ ($X \leftarrow B \rightarrow Y$).}
    \label{scm}
    \vspace{-1em}
\end{figure}

On the contrary, human cognitive systems are known to be immune to input perturbations such as stealthy trigger patterns induced by backdoor attacks \cite{gopnik2004theory}.
This is because humans are more sensitive to causal relations than the statistical associations of nuisance factors  \cite{ lake2017building, tenenbaum2011grow}. In contrast, deep learning models that are trained to fit the poisoned datasets can hardly distinguish the causal relations and the statistical associations brought by backdoor attacks. Based on causal reasoning, we can identify causal relation \cite{causality, causalinference} and build robust deep learning models \cite{zhang2020causal, zhang2022adversarial}. Therefore, it is essential to leverage causal reasoning to analyze and mitigate the threats of backdoor attacks.

In this paper, we focus on the image classification tasks and aim to train backdoor-free models on poisoned datasets without extra clean data. We first construct a causal graph to model the generation process of backdoor data where nuisance factors (\emph{i.e.,} backdoor trigger patterns) are considered. With the assistance of the causal graph, we find that the backdoor attack acts as the \emph{confounder} and opens a spurious path between the input image and the predicted label (Figure \ref{scm}). If DNNs have learned the correlation of such a spurious path, their predictions will be changed to the target labels when the trigger is attached.

Motivated by our causal insight, we propose {\bf C}ausality-inspired {\bf B}ackdoor {\bf D}efense (CBD) to learn deconfounded representations for classification. 
As the backdoor attack is stealthy and hardly measurable, we cannot directly block the backdoor path by the backdoor adjustment from causal inference \cite{causalinference}. Inspired by recent advances in disentangled representation learning \cite{wang2020cross, hamaguchi2019rare, liu2021mitigating}, we instead aim to learn deconfounded representations that only preserve the causality-related information. Specifically in CBD, two DNNs are trained, which focus on the spurious correlations and the causal effects respectively. The first DNN is designed to intentionally capture the backdoor correlations with an early stop strategy. The other clean model is then trained to be independent of the first model in the hidden space by minimizing mutual information. The information bottleneck strategy and sample-wise re-weighting scheme are also employed to help the clean model capture the causal effects while relinquishing the confounding factors.
After training, only the clean model is used for downstream classification tasks.
In summary, our contributions are:
\begin{itemize}
    \item From a causal perspective, we find the backdoor attack acts as the confounder that causes spurious correlations between the input images and the target label.
    \item With the causal insight, we propose a Causality-inspired Backdoor Defense (CBD), which learns deconfounded representations to mitigate the threat of poisoning-based backdoor attacks.
    \item Extensive experiments with 6 representative backdoor attacks are conducted. The models trained using CBD are of almost the same clean accuracy as they were directly trained on clean data and the average backdoor attack success rates are reduced to around 1$\%$, which verifies the effectiveness of CBD.
    \item We explore one potential adaptive attack against CBD, which tries to make the backdoor attack stealthier by adversarial training. Experiments show that CBD is robust and resistant to such an adaptive attack.
\end{itemize}

\section{Related Work}
\subsection{Backdoor Attacks}
Backdoor attacks are emerging security threats to deep neural network classifiers. In this paper, we focus on the poisoning-based backdoor attacks, where the attacker can only inject poisoned examples into the training set while cannot modify other training components (\emph{e.g.,} training loss). Note that backdoor attacks could occur in other tasks (\emph{e.g.,} visual object tracking~\cite{li2022few}, graph classification~\cite{zhang2021backdoor}, federated learning~\cite{zhang2022fldetector}, and multi-modal contrastive learning~\cite{backdoorcontrast}). The attacker may also have extra capabilities such as modifying the training process~\cite{latentbackdoor}. However, these situations are out of the scope of this paper.

Based on the property of target labels, existing backdoor attacks can be divided into two main categories: \emph{dirty-label attacks}~ \cite{gu2017badnets,chen2017targeted,liu2020reflection, zhang2021backdoor} and \emph{clean-label attacks}~\cite{shafahi2018poison,turner2019clean,zhu2019transferable,zhao2020clean,saha2020hidden}. Dirty-label attack is the most common backdoor attack paradigm, where the poisoned samples are generated by stamping the backdoor trigger onto the original images and altering the labels to the target label. BadNets~\cite{gu2017badnets} firstly employed a black-white checkerboard as the trigger pattern. Furthermore, more complex and stealthier trigger patterns are proposed such as blending backgrounds~\cite{chen2017targeted}, natural reflections~\cite{liu2020reflection}, invisible noise~\cite{liao2018backdoor,li2019invisible,chen2019invisible}  and sample-wise dynamic patterns \cite{nguyen2020input,li2021invisible, nguyen2021wanet}.
On the other hand, Clean-label backdoor attacks are arguably stealthier as they do not change the labels. For example, Turner \emph{et al.} \cite{turner2019clean} leveraged deep generative models to modify benign images from the target class.

\subsection{Backdoor Defenses}
Based on the target and the working stage, existing defenses can be divided into the following categories: (1) \emph{Detection based defenses} aim to detect anomalies in input data \cite{chen2018detecting,tran2018spectral,gao2019strip,xu2019detecting,hayase2021spectre,tang2021demon, Februus20} or whether a given model is backdoored \cite{wang2019neural,chen2019deepinspect,kolouri2020universal,shen2021backdoor}. For instance, Du \emph{et al.}~\cite{DPdetection} applies differential privacy to improve the utility of backdoor detection.
(2) \emph{Model reconstruction based defenses} aim to remove backdoors from a given poisoned model. For example, Mode Connectivity Repair (MCR)~\cite{zhao2020bridging} mitigates the backdoors by selecting a robust model in the path of loss landscape, while Neural Attention Distillation (NAD)~\cite{li2021neural} leverages attention distillation to remove triggers. (3) \emph{Poison suppression based defenses} \cite{li2021anti,backdoordecouple} reduce the effectiveness of poisoned examples at the training stage and try to learn a clean model from a poisoned dataset. For instance, Decoupling-based backdoor defense (DBD) \cite{backdoordecouple} decouples the training of DNN backbone and fully-connected layers to reduce the correlation between triggers and target labels. 
Anti-Backdoor Learning (ABL)~\cite{li2021anti} uses gradient ascent to unlearn the backdoored model with the isolated backdoor data. 

In this paper, our proposed CBD is most related to the \emph{Poison suppression based defenses}. Our goal is to train clean models directly on poisoned datasets without access to clean datasets or further altering the trained model. 
Different with ABL~\cite{li2021anti}, CBD directly trains clean models on poisoned datasets without further finetuning the trained model (unlearning backdoor). In contrast with DBD, CBD does not require additional self-supervised pretraining stages and is much more efficient. In Sec. \ref{sec: experiments}, extensive experimental results clearly show the advantages of CBD. 

\subsection{Causal Inference}
Causal inference has a long history in statistical research \cite{causality,bookwhy, causalinference}. The objective of causal inference is to analyze the causal effects among variables and mitigate spurious correlations. Recently, causal inference has also shown promising results in various areas of machine learning \cite{niu2021counterfactual,mitrovic2020representation, yue2020interventional,tang2020long,zhang2020causal,zhang2022adversarial, huang2022deconfounded}. 
However, to date, causal inference has not been incorporated into the analysis and defense of backdoor attacks.
\section{Preliminaries}
\subsection{Problem Formulation}
In this section, we first formulate the problem of poison suppression based defense, then provide a causal view on backdoor attacks and introduce our proposed Causality-inspired Backdoor Defense. Here, we focus on image classification tasks with deep neural networks.

\noindent\textbf{Threat Model. }We follow the attack settings in previous works \cite{li2021anti, backdoordecouple}. Specifically, we assume a set of backdoor examples has been pre-generated by the attacker and has been successfully injected into the training dataset. We also assume the defender has complete control over the training process but is ignorant of the distribution or the proportion of the backdoor examples in a given dataset. The defender’s goal is to train a backdoor-free model on the poisoned dataset, which is as good as models trained on purely clean data. Robust learning strategies developed under such a threat model could benefit research institutes, companies, or government agencies that have the computational resources to train their models but rely on outsourced training data.

\subsection{A Causal View on Backdoor Attacks}
Humans' ability to perform causal reasoning is arguably one of the most important characteristics that distinguish human learning from deep learning \cite{zhang2020causal, scholkopf2021toward}. The superiority of causal reasoning endows humans with the ability to recognize causal relationships while ignoring non-essential factors in tasks. On the contrary, DNNs usually fail to distinguish causal relations and statistical associations and tend to learn ``easier" correlations than the desired knowledge \cite{nam2020learning, geirhos2020shortcut}. Such a shortcut solution could lead to overfitting to nuisance factors (\emph{e.g.,} trigger patterns), which would further result in the vulnerability to backdoor attacks. Therefore, here we leverage causal inference to analyze DNN model training and mitigate the risks of backdoor injection. 

We first construct a causal graph as causal graphs are the keys to formalize causal inference. One
approach is to use causal structure learning to infer causal graphs \cite{causality},
but it is challenging to apply this kind of approach to high-dimensional data like images. Here, following previous works \cite{zhang2020causal, zhang2022adversarial, niu2021counterfactual}, we leverage domain knowledge (Figure \ref{scm} (a)) to construct a causal graph $\mathcal{G}$ (Figure \ref{scm} (b)) to model the generation process of poisoned data.

In the causal graph, we denote the abstract data variables by the nodes ($X$ as the input image, $Y$ as the label, and $B$ as the backdoor attack), and the directed links represent their relationships. As shown in Figure \ref{scm}(b), besides the causal effect of $X$ on $Y$ ($X \rightarrow Y$), the backdoor attacker can attach trigger patterns to images ($B \rightarrow X$) and change the labels to the targeted label ($B \rightarrow Y$). Therefore, as a \emph{confounder} between $X$ and $Y$, backdoor attack $B$ opens the spurious path $X \leftarrow B \rightarrow Y$ (let $B=1$ denotes the images are poisoned and $B=0$ denotes the images are clean). By \emph{“spurious”}, we mean that the path lies outside the direct causal path from $X$ to $Y$, making $X$ and $Y$ spuriously correlated and yielding an erroneous effect when the trigger is activated. DNNs can hardly distinguish between the spurious correlations and causative relations \cite{bookwhy}. Hence, directly training DNNs on potentially poisoned dataset incurs the risk of being backdoored. 

To pursue the causal effect of $X$ on $Y$, previous works usually perform the backdoor adjustment in the causal intervention \cite{bookwhy} with $do$-calculus: $P(Y |do(X)) = \sum_{B\in \{0,1\}} P(Y|X,B)P(B)$. 
However, since the confounder variable $B$ is hardly detectable and measurable in our setting, we can not simply use the backdoor adjustment to block the backdoor path. 
Instead, since the goal of most deep learning models is to learn accurate embedding representations for downstream tasks \cite{alemi2016deep, huang2021disenqnet, wang2020cross, hamaguchi2019rare}, we aim to disentangle the confouding effects and causal effects in the hidden space. The following section illustrates our method.

\section{Causality-inspired Backdoor Defense}
\begin{figure}[t]
    \centering
    \includegraphics[width=0.65\linewidth]{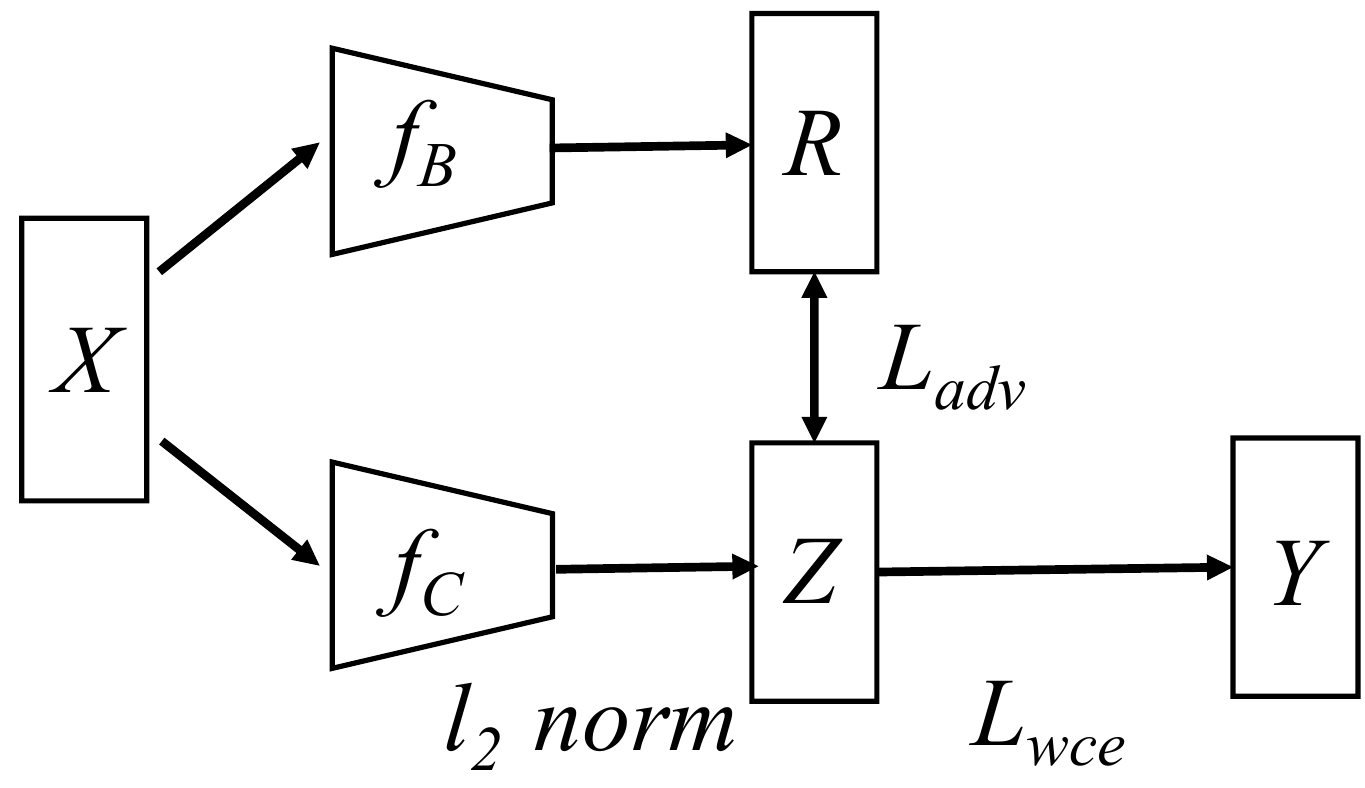}
    \caption{The model framework of CBD that includes an adversaral loss $\mathcal{L}_{adv}$ for mutual information minimization, a $l_2-$norm regularization on $z$, and a weighted cross entropy loss $\mathcal{L}_{wce}$ to augment causal effects.}
    \label{scm2}
    \vspace{-1em}
\end{figure}

Motivated by our causal insight, we propose the Causality-inspired Backdoor Defense (CBD). 
In practice, it may be difficult to directly identify the confounding and causal factors of $X$ in the data space. We make an assumption that the confounding and causal factors will be reflected in the hidden representations. 
Based on the assumption, we illustrate our main idea in Figure \ref{scm2}.
Generally, two DNNs including $f_B$ and $f_C$ are trained, which focus on the spurious correlations and the causal relations respectively. We take the embedding vectors from the penultimate layers of $f_B$ and $f_C$ as $R$ and $Z$. 
\emph{Note that $R$ is introduced to avoid confusion with $B$}.
Without confusion, we use \emph{uppercase} letters for variables and \emph{lowercase} letters for concrete values in this paper. To generate high-quality variable $Z$ that captures the causal relations, we get inspiration from disentangled representation learning \cite{wang2020cross, hamaguchi2019rare}.
In the training phase, $f_B$ is firstly trained on the poisoned dataset to capture spurious correlations of backdoor. The other clean model $f_C$ is then trained to encourage independence in the hidden space \emph{i.e.,} $Z \perp R$ with mutual information minimization and sample re-weighting. After training, only $f_C$ is used for downstream classification tasks. In the rest of this section, we provide details on each step of CBD. 

\noindent\textbf{Training a backdoored model $f_B$. }Firstly, $f_B$ is trained on the poisoned dataset with cross entropy loss to capture the spurious correlations of backdoor. Since the poisoned data still contains causal relations, we intentionally strengthen the confounding bias in $f_B$ with an \emph{early stop strategy}. Specifically, we train $f_B$ only for several epochs (\emph{e.g.,} 5 epochs) and freeze its parameters in the training of $f_C$. This is because previous works indicate that backdoor associations are easier to learn than causal relations \cite{li2021anti}. 
Experiments in Appendix \ref{app:more results} also verify that the losses on backdoor examples reach nearly 0 while $f_B$ has not converged on clean samples after 5 epochs.

\noindent\textbf{Training a clean model $f_C$.}
Inspired by previous works \cite{guo2019learning, huang2021disenqnet}, we formulate the training objective of $f_C$ with information bottleneck and mutual information minimization:
\begin{equation}
    \mathcal{L}_C={\rm min}\underbrace{\beta I(Z; X)}_{\circled{1}}- \underbrace{I(Z; Y)}_{\circled{2}}+\underbrace{I(Z; R)}_{\circled{3}},
    \vspace{-1em}
\label{lc}
\end{equation}
where $I(\cdot; \cdot)$ denotes the mutual information. Term $\circled{1}$ and $\circled{2}$ constitute the information bottleneck loss \cite{tishby2000information} that encourages the variable $Z$ to capture the core information for label prediction ($\circled{2}$) while constraining unrelated information from inputs ($\circled{1}$). $\beta$ is a weight hyper-parameter. Term $\circled{3}$ is a de-confounder penalty term, which describes the dependency degree between the backdoored embedding $R$ and the deconfounded embedding $Z$. It encourages $Z$ to be independent of $R$ by minimizing mutual information so as to focus on causal effects.
However, $\mathcal{L}_C$ in Equation \ref{lc} is not directly tractable, especially for the de-confounder penalty term. 
In practice, we relax Equation \ref{lc} and optimize the upper bound of the term $\circled{1}\&\circled{2}$ and the estimation of the term $\circled{3}$. The details are shown below.

\noindent \textbf{Term $\circled{1}$.} Based on the definition of mutual information and basics of probability, $I(Z; X)$ can be calculated as:
\begin{equation}
\begin{aligned}
    I(Z; X)&=\sum_x \sum_z p(z,x) {\rm log} \frac{p(z,x)}{p(z)p(x)} \\
    &=\sum_x \sum_z p(z|x)p(x) {\rm log} \frac{p(z|x)p(x)}{p(z)p(x)}\\
    &=\sum_x \sum_z p(z|x)p(x){\rm log}~ p(z|x)-\sum_z p(z){\rm log}~p(z).
    \label{I(z;x)}
\end{aligned}
\end{equation}
However, the marginal probability $p(z) = \sum_x p(z|x)p(x)$ is usually difficult to calculate in practice. We use variational approximation to address this issue, \emph{i.e.,} we use a variational distribution $q(z)$ to approximate $p(z)$. According to Gibbs’ inequality \cite{mackay2003information}, we know that the KL divergence is non-negative: $D_{\rm KL}(p(z)||q(z))\ge 0 \Rightarrow -\sum_z p(z) {\rm log}~p(z)\le -\sum_z~p(z) {\rm log}~q(z)$. By substitute such inequality into Equation \ref{I(z;x)}, we can derive an upper bound of $I(Z;X)$:
\begin{equation}
\begin{aligned}
    I(Z; X)&\le\sum_x p(x)\sum_z p(z|x){\rm log}~ p(z|x)-\sum_z p(z){\rm log}~q(z)\\
    &=\sum_x p(x)\sum_z p(z|x) {\rm log} \frac{p(z|x)}{q(z)}\\
    &=\sum_x p(x)D_{\rm KL}(p(z|x)||q(z)).
    \label{upper bound}
\end{aligned}
\end{equation}
Following previous work \cite{alemi2016deep}, we assume that the posterior $p(z|x)=\mathcal{N}(\mu(x),{\rm diag}\{\sigma^2(x)\})$ is a gaussian distribution, where $\mu(x)$ is the encoded embedding of variable $x$ and ${\rm diag} \{\sigma^2(x)\} = \{\sigma_d^2\}_{d=1}^D$ is the diagonal matrix indicating the variance. On the other hand, the prior $q (z)$ is assumed to be a standard Gaussian distribution, \emph{i.e.,} $q(z) = \mathcal{N}(0, I)$. Finally, we can rewrite the above upper bound as:
\begin{equation}
D_{\rm KL}(p(z|x)||q(z))=\frac{1}{2}||\mu(x)||_2^2+\frac{1}{2}\sum_d(\sigma_d^2- {\rm log}\sigma_d^2 -1).
\label{kl gaussian}
\end{equation}
The detailed derivation is shown in Appendix \ref{app: derivation}.
For ease of optimization, we fix $\sigma(x)$ to be an all-zero matrix.  
Then $z = \mu(x)$ becomes a deterministic embedding. The optimization of this upper bound is equivalent to directly applying the $l_2$-norm regularization on the embedding vector $z$.

\noindent \textbf{Term $\circled{2}$.} With the definition of mutual information, we have $I (Z;Y) = H (Y)-H (Y|Z)$, where $H(\cdot)$ and $H(\cdot|\cdot)$ denote the entropy and conditional
entropy respectively. Since $H (Y)$ is a positive constant and can be ignored, we have the following inequality,
\begin{equation}
    -I(Z;Y)\le  H(Y|Z).
\end{equation}
In experiments, $H(Y|Z)$ can be calculated and optimized as the cross entropy loss ($CE$). To further encourage the independence between $f_C$ and $f_B$,
we fix the parameters of $f_B$ and train $f_C$ with the samples-wise weighted cross entropy loss ($\mathcal{L}_{wce}$). The weight is calculated as:
\begin{equation}
    w(x) = \frac{CE(f_B(x), y)}{CE(f_B(x),y)+CE(f_C(x),y)}.
\end{equation}
For samples with large losses on $f_B$, $w(x)$ are close to 1; while $w(x)$ are close to 0 when the losses are very small. The intuition of the re-weighting scheme is to let $f_C$ focus on ``hard" examples for $f_B$ to encourage independence. 

\noindent \textbf{Term $\circled{3}$.} Based on the relationship between mutual information and Kullback-Leibler (KL) divergence \cite{belghazi2018mutual}, the term $I(Z; R)$ is equivalent to the KL divergence between the joint distribution $p(Z, R)$ and the product of two marginals $p(Z)p(X)$ as: $I(Z; R)=D_{\rm KL}(p(Z, R)||p(Z) p(R))$. Therefore, to minimize the de-confounder penalty term $I(Z;R)$, we propose to use an adversarial process that minimizes the distance between the joint distribution $p(Z,R)$ and the marginals $p(Z) p(R)$. During the adversarial process, a discriminator $D_\phi$ is trained to
classify the sampled representations drawn from the joint $p(Z,R)$ as the real, \emph{i.e.,} 1 and samples drawn from the marginals $p(Z) p(R)$ as the fake, \emph{i.e.,} 0. 
The samples from the marginal distribution $p(Z) p(R)$ are obtained by shuffling the individual representations of samples $(z,r)$ in
a training batch from $p(Z,R)$. On the other hand, the clean model $f_C$ tries to generate $z$ that look like
drawn from $p(Z) p(R)$ when combined with $r$ from $f_B$. 
Specifically, we optimize such adversarial objective similar to WGAN \cite{arjovsky2017wasserstein} with spectral normalization \cite{miyato2018spectral} since it is more stable in the learning process:
\begin{equation}
    \mathcal{L}_{adv} =\mathop{\rm min}\limits_{\theta_C} \mathop{ \rm max}\limits_{\phi} \mathbb{E}_{p(z,r)}[D_\phi(z,r)] - \mathbb{E}_{p(z)p(r)}[D_\phi(z,r)],
\end{equation}
where $\theta_C$ and $\phi$ denote the parameters of $f_C$ and $D_\phi$ respectively. To sum up, the training objective for $f_C$ is:
\begin{equation}
    \mathcal{L}_C = \mathcal{L}_{wce}+\mathcal{L}_{adv}+\beta||\mu(x)||_2^2.
\label{loss function}
\end{equation}
The overall objective can then be minimized using SGD. $f_B$ and $f_C$ are trained for $T_1$ and $T_2$ epochs respectively. The pseudo algorithm of CBD is shown in Algorithm \ref{algor}. 
\begin{algorithm}[t]
\caption{Causality-inspried Backdoor Defense (CBD)}
\textbf{Input}: $\beta$, number of training iterations $T_1, T_2$\\
\textbf{Output}: Clean model $f_C$;
\begin{algorithmic}[1]
\STATE Initialize $f_C$, $f_B$, and $D_\phi$\\
\FOR{$t= 1, \cdots, T_1$}
\STATE Train $f_B$ on the poisoned dataset with SGD
\ENDFOR
\FOR{$t= 1, \cdots, T_2$}
\STATE Train discriminator $D_\phi$
\STATE Calculate sample weight $w(x)$
\STATE Train $f_C$ with loss function in Equation \ref{loss function}
\ENDFOR
\end{algorithmic}\label{algor}
\end{algorithm}

\noindent\textbf{Further Discussions.} 
Admittedly, it is challenging to disentangle causal factors and confounding factors thoroughly. This is because $f_B$ may still capture some causal relations. Moreover, encouraging independence between $Z$ and $R$ may result in loss of predictive information for $f_C$.
However, with the well-designed optimization objectives and training scheme, CBD manages to reduce the confounding effects as much as possible while preserving causal relations. The following section shows the detailed verification.

\begin{table*}[t]
\renewcommand{\arraystretch}{1.1}
\renewcommand\tabcolsep{1.6pt}
\small
\centering
\caption{The attack success rate (ASR \%) and the clean accuracy (CA \%)  of 5 backdoor defense methods against 6 representative backdoor attacks. \emph{None} means the training data is completely clean. The best results are bolded.}
  \label{tab1}
\begin{tabular}{c|c|cc|cc|cc|cc|cc|cc|cc}
\toprule
\multirow{2}{*}{Dataset} & \multirow{2}{*}{Types} & \multicolumn{2}{c|}{\begin{tabular}[c|]{@{}c@{}}No Defense\end{tabular}} & \multicolumn{2}{c|}{FP} & \multicolumn{2}{c|}{MCR} & \multicolumn{2}{c|}{NAD} & \multicolumn{2}{c|}{ABL}&\multicolumn{2}{c|}{DBD}& \multicolumn{2}{c}{\textbf{CBD (Ours)}}\\ \cline{3-16} 
 &  & ASR & CA & ASR & CA & ASR & CA & ASR & CA & ASR & CA& ASR & CA& ASR & CA\\ \hline
\multirow{8}{*}{CIFAR-10} 
& \emph{None} & 0 & 89.14 & 0 & 85.17 & 0 & 87.55 & 0 & 88.21 & 0 & 88.49 &0 & 88.63 &0 & \bf 88.95 \\ \cline{2-16}
& BadNets & 100 & 85.37 & 99.96 & 82.41 & 4.52 & 79.66 & 3.07 & 82.25 & 3.13 & 86.30 & 1.76&86.94&\bf 1.06 &\bf 87.46\\
 & Trojan & 100 & 84.54 & 68.95 & 81.03 & 19.47 & 77.12 & 19.96 & 80.05 & 3.88 & 87.29&3.79&87.29& \bf1.24 & \bf87.52 \\
 & Blend & 100 & 84.56 & 87.14 & 81.57 & 36.15 & 78.24 & 10.65 & 83.71 & 14.60 & 85.02 &5.12&86.83& \bf 1.96&\bf 87.48\\
 & SIG & 99.32 & 84.14 & 73.87 & 81.04 & 2.34 & 77.93 & 1.79 & 83.54 & 0.36 &\bf88.10 &0.44&87.52& \bf0.25 & 87.29 \\
 & Dynamic & 100 & 85.48 & 89.22 & 80.63 & 25.26 & 75.03 & 25.60 & 74.94 & 17.26 & 85.29 &10.21&85.42&\bf 0.86& \bf 85.67 \\
 &  WaNet & 98.55 & 86.77 & 73.12 & 81.58 & 28.59& 77.12 & 24.15 & 79.50 & 22.24 & 75.74 &5.86&84.60&\bf 4.24&\bf 86.55 \\\cline{2-16} 
 & \multicolumn{1}{l}{Average} & \multicolumn{1}{|l}{99.65} & \multicolumn{1}{l}{85.14} & \multicolumn{1}{|l}{82.03} & \multicolumn{1}{l}{81.38} & \multicolumn{1}{|l}{19.39} & \multicolumn{1}{l}{77.52} & \multicolumn{1}{|l}{14.20} & \multicolumn{1}{l}{80.67}  & \multicolumn{1}{|c}{10.25} & \multicolumn{1}{c}{84.62}&\multicolumn{1}{|c}{4.53}&\multicolumn{1}{c}{86.43}& \multicolumn{1}{|c}{\bf 1.60} & \multicolumn{1}{c}{\bf 87.00} \\ \midrule
\multirow{8}{*}{GTSRB} 
& \emph{None} & 0 & 97.74 & 0 & 90.18 & 0 & 95.27 & 0 & 95.29 & 0 & 96.47&0&96.45& 0 & \bf 96.54 \\ \cline{2-16} 
& BadNets & 100 & 96.58 & 99.48 & 88.57 & 1.27 & 93.30 & 0.31 & 89.90 & \textbf{0.05}&96.01&0.24&96.05& 0.16 & \bf 96.21\\
 & Trojan & 99.95 & 96.49 & 97.40 & 88.51 & 4.62 & 92.99 & 0.56 & 90.32 & 0.47 & 94.91&0.56&94.69&\bf 0.12 & \bf 95.29 \\
 & Blend & 100 & 95.57 & 98.78 & 87.50 & 6.85 & 93.11 & 13.06 & 89.20 & 22.97 & 93.25 &6.36&93.72& \bf 0.90 & \bf 94.16 \\
 & SIG & 98.24 & 96.55 & 85.04 & 89.97 & 26.80 & 91.14 & 5.35 & 89.27 & 5.09 &\bf96.28 &\bf4.70&94.58& 5.41 &  94.37\\
 & Dynamic & 100 & 96.87 & 98.33 & 88.09 & 59.54 & 90.51 & 62.35 & 84.30 & 6.32 & 95.76 &5.16&95.86&\bf0.96&\bf96.02 \\
  & WaNet & 99.92 & 95.94 & 97.93 & 90.13 & 55.25 & 91.24 & 34.16 & 83.09 & 5.56 & 93.83&3.47&94.71 &\bf 3.13 &\bf95.64 \\
  \cline{2-16} 
 & \multicolumn{1}{l}{Average} & \multicolumn{1}{|l}{99.69} & \multicolumn{1}{l}{96.33} & \multicolumn{1}{|l}{96.16} & \multicolumn{1}{l}{88.80} & \multicolumn{1}{|l}{25.72} & \multicolumn{1}{l}{92.05} & \multicolumn{1}{|l}{19.30} & \multicolumn{1}{c}{87.68} &  \multicolumn{1}{|c}{7.96} & \multicolumn{1}{c}{95.01}&\multicolumn{1}{|c}{3.42}&\multicolumn{1}{c}{94.94}& \multicolumn{1}{|c}{\textbf{1.82}} & \multicolumn{1}{c}{\textbf{95.17}}\\ \midrule
\multirow{6}{*}{\begin{tabular}[c]{@{}c@{}}ImageNet\\ Subset\end{tabular}} 
& \emph{None} & 0 & 88.95 & 0 & 83.05 & 0 & 85.61 & 0 & 87.34 & 0 & 88.12 &0&88.30&0&\bf88.57 \\ \cline{2-16} 
& BadNets & 100 & 85.24 & 98.03 & 82.76 & 25.14 & 77.90 & 7.38 & 82.11 & 1.02 & 87.47&1.27&87.61& \bf0.66&\bf88.12 \\
 & Trojan & 100 & 85.65 & 97.29 & 81.46 & 6.65 & 77.06 & 13.80 & 81.49 & 1.68 & 88.21&1.48&88.20&\bf 0.72& \bf88.24 \\
 & Blend & 99.89 & 86.10 & 99.10 & 81.37 & 18.37 & 76.21 & 25.05 & 82.54 & 20.80 & 85.23&4.73&86.25&\bf1.82& \bf87.95 \\ 
 & SIG & 98.53 & 86.06 & 77.39 & 82.55 & 24.62 & 78.97 & 5.30 & 83.24 & \textbf{0.22} & 86.65 &1.95&87.09&0.45&\bf87.27\\
\cline{2-16} 
 & \multicolumn{1}{l}{Average} & \multicolumn{1}{|l}{99.61} & \multicolumn{1}{l}{85.74} & \multicolumn{1}{|l}{92.95} & \multicolumn{1}{l}{82.04} & \multicolumn{1}{|l}{18.70} & \multicolumn{1}{l}{77.54} & \multicolumn{1}{|l}{12.88} & \multicolumn{1}{l}{82.35} & \multicolumn{1}{|c}{5.93} & \multicolumn{1}{c}{86.89}& \multicolumn{1}{|c}{2.36} & \multicolumn{1}{c}{87.29} &\multicolumn{1}{|c}{\textbf{0.91}} & \multicolumn{1}{c}{\textbf{87.90}} \\
 \bottomrule
\end{tabular}
\label{core experiment}
\vspace{-0.1in}
\end{table*}

\begin{table*}[!tp]
\renewcommand{\arraystretch}{1.2}
\renewcommand\tabcolsep{1.8pt}
\small
\centering
\caption{Robustness test with the poisoning rate from 1\% to 50\% for 4 attacks including BadNets, Trojan, Blend, and WaNet on CIFAR10 dataset. We show ASR ($\%$) and CA ($\%$).}
\begin{tabular}{c|c|cccccccc}
\hline
\multirow{2}{*}{\begin{tabular}[c]{@{}c@{}}Poisoning\\ Rate\end{tabular}} & \multirow{2}{*}{Defense} & \multicolumn{2}{c}{BadNets} & \multicolumn{2}{c}{Trojan} & \multicolumn{2}{c}{Blend} &  \multicolumn{2}{c}{WaNet} \\ \cline{3-10} 
 &  & ASR & CA & ASR & CA & ASR & CA & ASR & CA \\ \hline
\multirow{2}{*}{1\%} & \emph{None} & 100 & 85.67 & 100 & 85.15 & 100 & 85.22 & 97.56 & 86.55 \\ 
 & CBD & 0.62 & 88.83 & 1.13 & 88.56 & 0.67 & 87.52 & 1.06 & 86.59 \\ \hline
 \multirow{2}{*}{5\%} & \emph{None} & 100 & 84.68 & 100 & 84.82 & 100 & 85.06 & 99.83 & 86.27 \\ 
 & CBD & 0.93 & 87.50 & 1.10 & 88.45 & 0.73 & 87.47 & 1.07 & 86.56 \\ \hline
 \multirow{2}{*}{20\%} & \emph{None} & 100 & 83.42 & 100 & 79.32 & 100 & 82.08 & 100 & 74.41 \\ 
 & CBD & 1.16 & 84.35 & 1.57 & 81.71 & 5.17 & 86.53 & 5.72 & 74.25 \\ \hline
\multirow{2}{*}{50\%} & \emph{None} & 100 & 79.45 & 100 & 72.83 & 100 & 69.67 & 100 & 67.25 \\  
 & CBD & 1.47 & 78.88 & 2.31 & 75.34 & 8.14 & 85.56 & 8.75 & 70.43 \\ \hline
\end{tabular}
\vspace{-0.1in}
\label{poison rate}
\end{table*}

\section{Experiments}
\label{sec: experiments}
\subsection{Experimental Settings}
\label{experiment settings}
\noindent\textbf{Datasets and DNNs.} 
We evaluate all defenses on three classical benchmark datasets, CIFAR-10 \cite{krizhevsky2009learning}, GTSRB \cite{stallkamp2012man} and an ImageNet subset \cite{deng2009imagenet}. As for model architectures, we adopt WideResNet (WRN-16-1) \cite{zagoruyko2016wide} for CIFAR-10 and GTSRB and ResNet-34 \cite{he2016deep} for ImageNet subset. Note that our CBD is agnostic to the model architectures. Results with more models are shown in Appendix \ref{app:more results}.

\noindent\textbf{Attack Baselines.} 
We consider 6 representative backdoor attacks. Specifically, we select
BadNets~\cite{gu2017badnets}, Trojan attack~\cite{liu2018trojaning}, Blend attack~\cite{chen2017targeted}, Sinusoidal signal attack~(SIG)~\cite{barni2019new}, Dynamic attack~\cite{nguyen2020input}, and WaNet~\cite{nguyen2021wanet}.  BadNets, Trojan attack are patch-based visible dirty-label attacks; Blend is an invisible dirty-label attack; SIG belongs to clean-label attacks; Dynamic and WaNet are dynamic dirty-label attacks. More types of backdoor attacks are explored in Appendix \ref{app:more results}. The results when there is no attack and the dataset is completely clean is shown for reference.

\noindent\textbf{Defense Baselines.}
We compare our CBD with 5 state-of-the-art defense methods: Fine-pruning (FP)~\cite{liu2018fine}, Mode Connectivity Repair (MCR)~\cite{zhao2020bridging}, Neural Attention Distillation (NAD)~\cite{li2021neural}, Anti-Backdoor Learning (ABL)~\cite{li2021anti}, and Decoupling-based backdoor defense (DBD) \cite{backdoordecouple}. We also provide results of DNNs trained without any defense methods \emph{i.e.,} No Defense.

\noindent\textbf{Attack Setups.}
We trained DNNs on poisoned datasets for 100 epochs using Stochastic Gradient Descent (SGD) with an initial learning rate of 0.1 on CIFAR-10 and the ImageNet subset (0.01 on GTSRB), a weight decay of 0.0001, and a momentum of 0.9. The target labels of backdoor attacks are set to 0 for CIFAR-10 and ImageNet, and 1 for GTSRB. The default poisoning rate is set to 10$\%$.

\noindent\textbf{Defense Setups.}
For FP, MCR, NAD, ABL, and DBD, we follow the settings specified in their original papers, including the available clean data. 
Three data augmentation techniques suggested in \cite{li2021anti} including random crop, horizontal flipping, and cutout, are applied for all defense methods. The hyper-parameter $T_1$ and $\beta$ are searched in $\{3,5,8\}$ and $\{1e^{-5}, 1e^{-4}, 1e^{-3}\}$ respectively. Following the suggestion of the previous work \cite{zhao2020bridging}, we choose hyperparameters with 5-fold cross-validation on the training set according to the average classification accuracy on hold-out sets. $T_2$ is set to 100 in the default setting. 
All experiments were run one NVIDIA Tesla V100 GPU.
More details of settings are shown in Appendix \ref{app:settings}. 

\noindent\textbf{Metrics.} We adopt two commonly used performance metrics for the evaluation all methods: \emph{attack success rate} (ASR) and \emph{clean accuracy} (CA).
Let $\mathcal{D}_{test}$ denotes the benign testing set and $f$ indicates the trained classifier, we have ${\rm ASR} \triangleq \Pr_{({x},y) \in \mathcal{D}_{test}}\{f(B({x}))=y_t|y \neq y_t\}$ and ${\rm CA} \triangleq \Pr_{({x},y) \in \mathcal{D}_{test}}\{f({x})=y\}$, where $y_t$ is the target label and $B(\cdot)$ is the adversarial generator to add triggers into images. 
Overall, the lower the ASR and the higher the BA, the better the defense.

\subsection{Effectiveness of CBD}
\noindent\textbf{Comparison to Existing Defenses.} Table \ref{core experiment} demonstrates the proposed CBD method on CIFAR-10,
GTSRB, and ImageNet Subset. We consider 6 representative backdoor attacks and compare the performance of CBD with four other backdoor defense techniques. We omit some attacks on ImageNet dataset due to a failure of reproduction following their original papers.
We can observe that CBD achieves the lowest average ASR across all three datasets. Specifically, the average ASRs are reduced to around 1$\%$ (1.60$\%$ for CIFAR-10, 1.82$\%$ for GTSRB, and 0.91$\%$ for ImageNet). 
On the other hand, the CAs of CBD are maintained and are close to training DNNs on clean datasets without attacks.
We argue that baseline methods which try to fine-prune or unlearn backdoors of backdoored models are sub-optimal and less efficient. For example, the ABL tries to unlearn backdoor after model being backdoored; DBD requires additional a self-supervised pre-training stage, which introduces around 4 times overhead \cite{backdoordecouple}. On the contrary, CBD directly trains a backdoor-free model, which achieves high clean accuracy while keeping efficiency. \label{effectiveness}

When comparing the performance of CBD against different backdoor attacks, we find WaNet achieves higher ASR than most attacks consistently (4.24$\%$ for CIFAR-10, 3.41$\%$ for GTSRB). This may be explained by the fact that WaNet  as one of the state-of-the-art backdoor attacks adopts image warping as triggers that are stealthier than patch-based backdoor attacks \cite{nguyen2021wanet}. Hence, the spurious correlations between backdoor triggers and the target label are more difficult to capture for $f_B$. Then in the second step, $f_C$ struggles to distinguish the causal and confounding effects. We also notice that CBD is not the best when defending SIG on GTSRB and ImageNet Subset. We guess the reason is similar to WaNet discussed above. SIG produces poisoned samples by mingling trigger patterns with the background. Moreover, SIG belongs to clean-label attacks, which are stealthier than dirty-label attacks \cite{barni2019new}. This is one limitation of our CBD to be improved in the future.

\noindent\textbf{Effectiveness with Different Poisoning Rate. }In Table \ref{poison rate}, we demonstrate that our CBD is robust and can achieve satisfactory defense performance with a poisoning rate ranging from 1$\%$ to 50$\%$. Note that the results when poisoning rate equals $0\%$ have been shown in Table \ref{core experiment} (\emph{None}). Here, we did experiments on CIFAR-10
against 4 attacks including BadNets, Trojan, Blend, and WaNet. Generally, with a higher poisoning rate, CBD has lower CA and higher ASR. We can find that even with a poisoning rate of up to 50$\%$, our CBD
method can still reduce the ASR from 100$\%$ to 1.47$\%$, 2.31$\%$, 8.14$\%$, and 8.75$\%$ for BadNets, Trojan, Blend, and WaNet, respectively. Moreover, CBD helps backdoored DNNs recover clean accuracies. For instance, the CA of CBD for Blend and WaNet improves from 69.67$\%$ and 67.25$\%$ to 85.56$\%$ and 70.43$\%$ respectively at 50$\%$ poisoning rate.

\begin{figure*}[t]
	\centering
	\subfloat[Initialization]{\includegraphics[width=0.24\linewidth]{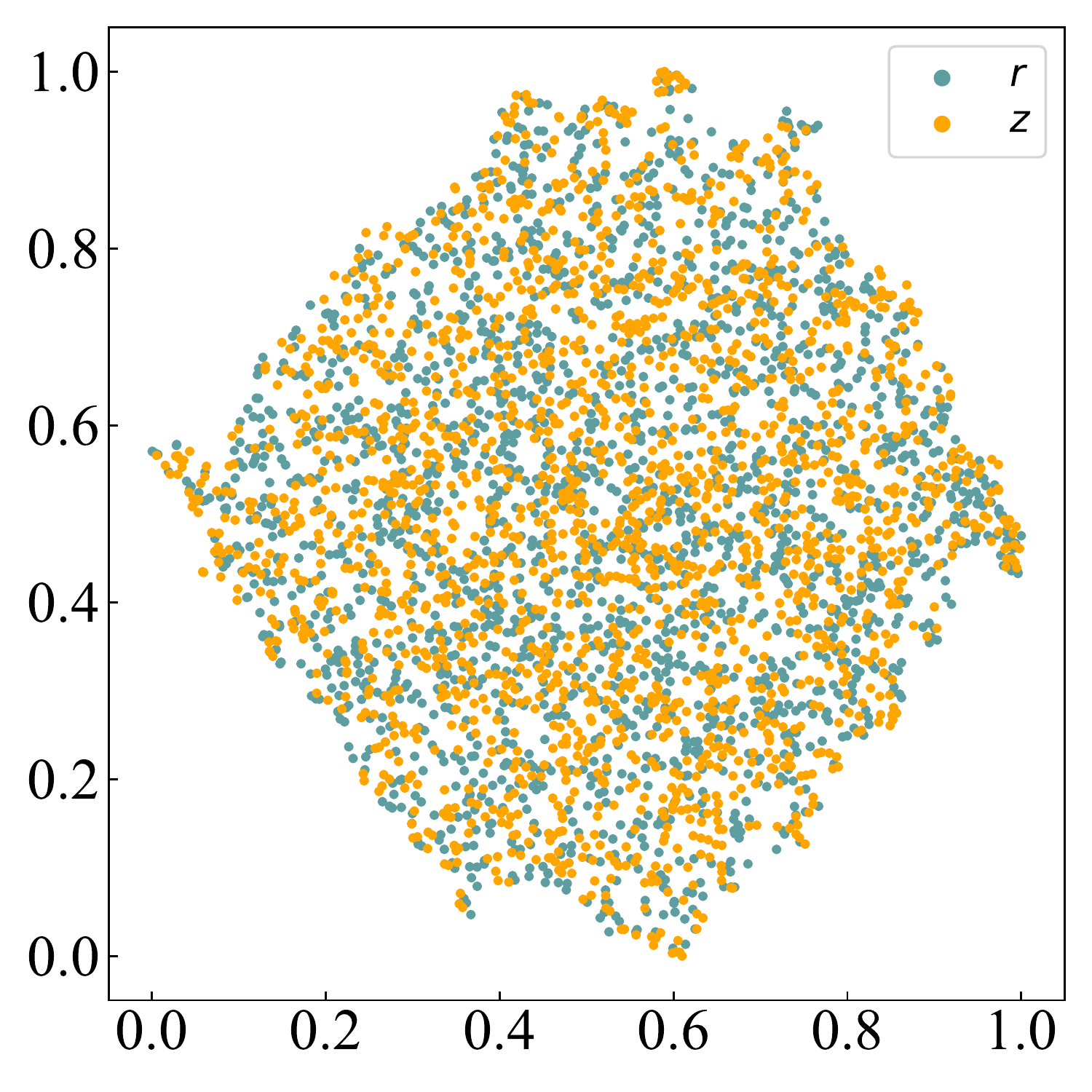}}
    \subfloat[Training Completed]{\includegraphics[width=0.24\linewidth]{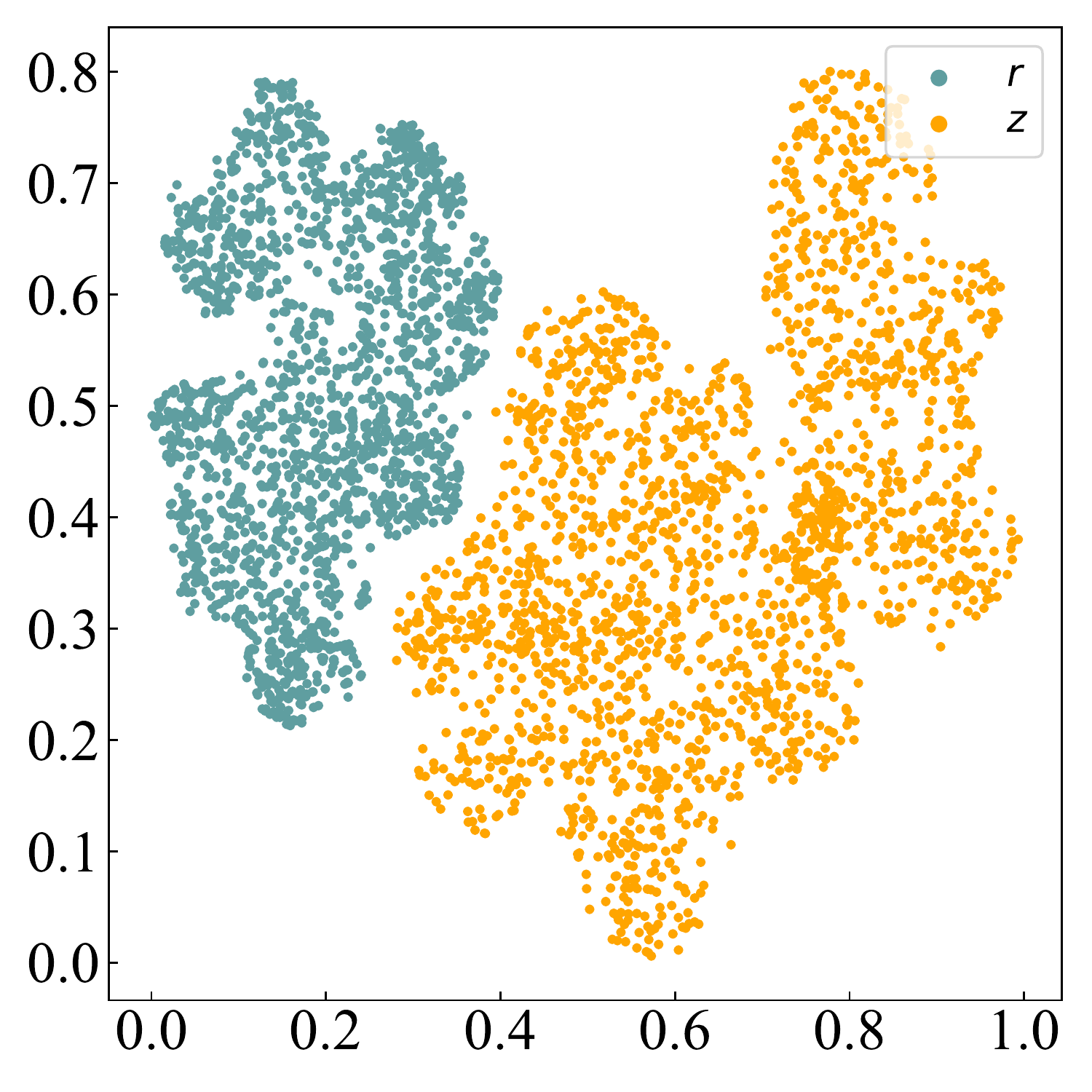}}
    \subfloat[$r$]{\includegraphics[width=0.24\linewidth]{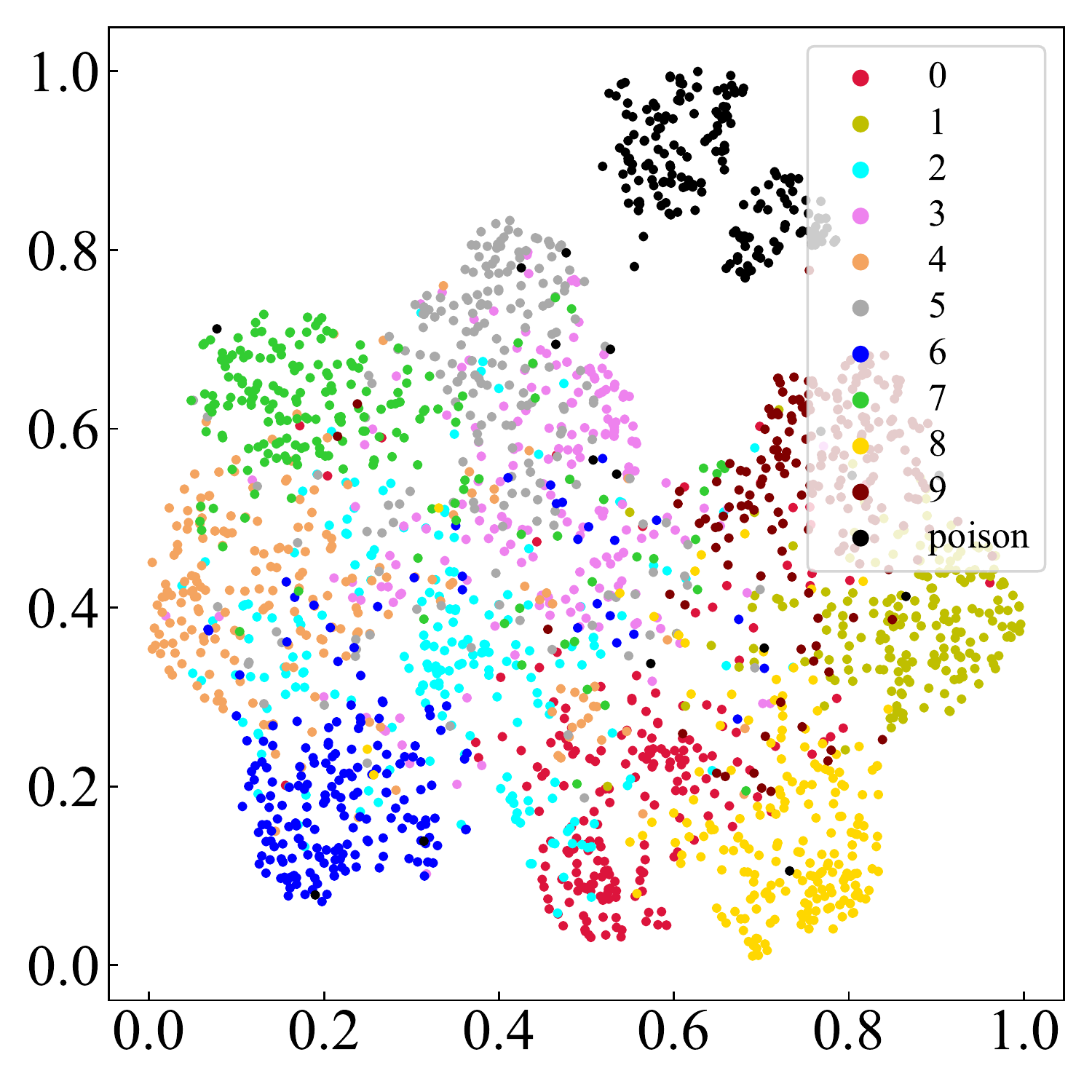}}
    \subfloat[$z$]{\includegraphics[width=0.24\linewidth]{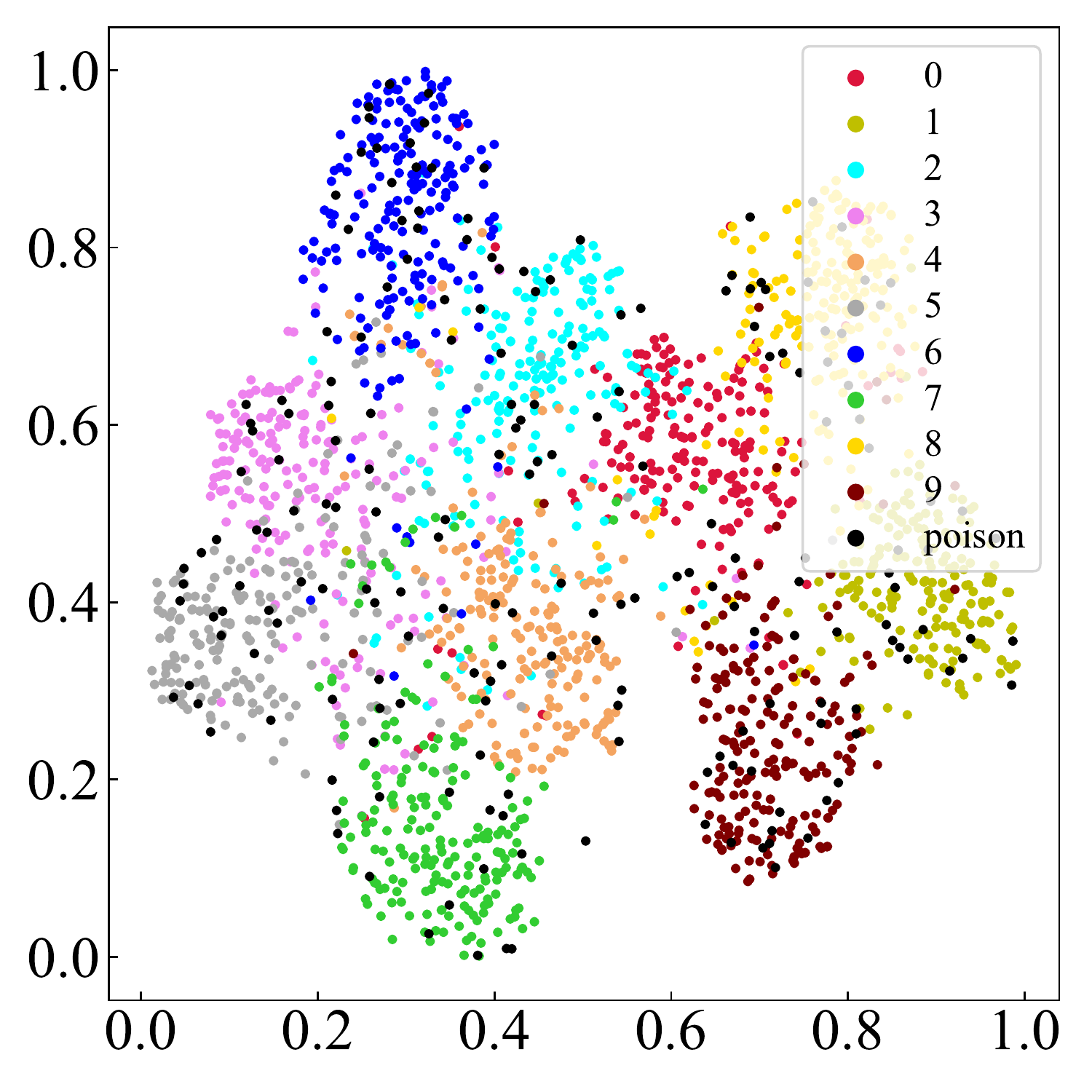}}
	\caption{Visualization of the hidden space with t-SNE}
	\label{tsne}
\end{figure*}

\noindent \textbf{Visualization of the Hidden Space.}
In Figure \ref{tsne}, we show the t-SNE \cite{JMLR:v9:vandermaaten08a} visualizations of the embeddings to give more insights of our proposed method. We conduct the BadNets attack on CIFAR-10. First, in Figure \ref{tsne} (a)$\&$(b), we show the embeddings of $r$ and $z$ when CBD is just initialized and when the training of CBD is completed. We observe that there is a clear separation between the confounding component $r$ and the causal component $z$ after training. Moreover, in Figure \ref{tsne} (c)$\&$(d), we use t-SNE separately on $r$ and $z$ and mark samples with different labels with different colors. Interestingly, we find the embeddings of the poisoned samples form clusters in $r$, which indicates that the spurious correlation between backdoor trigger and the target label has been learned. In contrast, poisoned
samples lie closely to samples with their ground-truth label in deconfounded embeddings $z$, which demonstrates CBD can effectively defend backdoor attacks. 

\noindent \textbf{Computational Complexity. }Compared with the vanilla SGD to train a DNN model, CBD only requires additionally training a backdoored model $f_B$ for a few epochs (\emph{e.g.,} 5 epochs) and a discriminator $D_\phi$, which introduces minimal extra overhead. 
Here, we report the training time cost of CBD on CIFAR-10 and the ImageNet
subset in Table \ref{time cost}. We also report the time costs of training vanilla DNNs for reference.
The extra computational cost is around $10\%-20\%$ of the standard training time on CIFAR-10 and the ImageNet subset.  This again shows the advantages of our method.

\begin{table}[t]
  \caption{The average training time (seconds) on CIFAR10 and the ImageNet subset with no defense and CBD. The percentages in parentheses indicate the relative increase of training time.}
\scriptsize
  \centering
  \begin{tabular}{c|cccc}
    \toprule
     \multirow{2}{*}{Dataset} & \multicolumn{2}{c}{CIFAR-10} & \multicolumn{2}{c}{ImageNet subset} \\\cline{2-5} 
 & No Defense & CBD & No Defense & CBD \\
    \hline
    BadNets& 1152& 1317{\tiny(14.3$\%$)}&2640 & 2987{\tiny(13.1$\%$)}\\
    Trojan& 1204& 1356{\tiny(12.6$\%$)}& 2621& 2933{\tiny(11.9$\%$)}\\
    Blend&1159 & 1311{\tiny(13.1$\%$)}& 2623& 3076{\tiny(17.3$\%$)}\\
    WaNet&1164& 1293{\tiny(11.1$\%$)}& 2647& 3074{\tiny(16.1$\%$)}\\ 
    \bottomrule
  \end{tabular}
  \label{time cost}
\end{table}
\subsection{Resistance to Potential Adaptive Attacks}\label{adaptive attack}
While not our initial intention, our work may be used to help develop more advanced backdoor attacks. Here, we tentatively discuss the potential adaptive attacks on CBD. Typically, backdoor attacks are designed to be injected successfully in a few epochs even only a small portion of data is poisoned (\emph{e.g.,} less than 1$\%$). Hence, the confounding bias of backdoor can be well captured by $f_B$ and $R$. 
The intuition of our adaptive attack strategy is to slow the injection process of backdoor attacks (\emph{i.e.,} increasing the corresponding training losses) by adding optimized noise into the poisoned examples, similar to recent works on adversarial training \cite{madry2017towards} and unlearnable examples \cite{huang2021unlearnable}.
If the confounding effects is not captured in the first step, then our CBD becomes ineffective. 

\noindent \textbf{Assumptions on Attacker's Capability. }
We assume that attackers have the entire benign dataset. The attackers may have the knowledge of the DNN architecture but cannot interfere with the training process. Moreover, the attackers cannot further modify their poisoned data once the poisoned examples are injected into the training dataset.

\noindent \textbf{Problem Formulation.} We formulate the adaptive attack as a min-max optimization problem. Let $\mathcal{D}'=\{({x}_i, y_i)\}_{i=1}^{N'}$ indicates the poisoned images by backdoor attacks,  $\mathcal{D}=\{({x}_i, y_i)\}_{i=1}^{N}$ denotes the benign images, and $\delta_i$ is the added noise to be optimized. Given an DNN model $f_{\theta}$ with parameters ${\theta}$, the adaptive attack aims to optimize $\delta_i$ by maximizing the losses of poisoned examples while minimizing the average cross entropy losses of all the samples, \emph{i.e.,}
\begin{equation}
    \min_{\theta}\left[\sum_{x \in \mathcal{D}} \mathcal{L}(f_\theta(x),y)+\sum_{x \in \mathcal{D}'} \max_{\delta_i}\mathcal{L}(f_\theta(x+\delta_i),y)\right],
\end{equation}
where the noise $\delta_i$ is bounded by $\|\delta_i\|_p \le \epsilon$ with $\|\cdot\|_p$ denoting the $L_p$ norm, and $\epsilon$ is set to be small such that the poisoned samples cannot be filtered by visual inspection. After optimization,  the poisoned examples attached with the optimized noises $\delta_i$ are injected to the training dataset.
We adopt the first-order optimization method PGD \cite{madry2017towards} to solve the constrained inner maximization problem:
\begin{equation}
    x_{t+1} = \Pi_\epsilon(x_t + \alpha\cdot \nabla_x\mathcal{L}(f_{\theta}(x_t), y)),
\end{equation}
where $t$ is the current perturbation step ($M$ steps in total), $\nabla_x\mathcal{L}(f_{\theta}(x_t), y)$ is the gradient of the loss
with respect to the input, $\Pi_\epsilon$ is a projection function that clips the noise back to the $\epsilon$-ball around
the original example $x$ when it goes beyond, and $\alpha$ is the step size. Pseudo codes of the adaptive attack are shown in Appendix \ref{app:more results}.

\noindent \textbf{Experimental Settings.}
We adopt the CIFAR-10 dataset and WRN-16-1 to conduct the experiments. According to previous studies in adversarial attacks, small $L_\infty$-bounded noise within $\|\delta\|_\infty \textless 8/255$ on images are unnoticeable to human observers. Therefore, we consider the same constraint in our experiments.
We use the SGD to solve the above optimization problem for 10 epochs with the step size $\alpha$ of 0.002 and $M = 5$ perturbation steps.

\noindent \textbf{Results.}
With BadNets, the adaptive attack works well when there is no defense (CA=84.55\%, ASR=99.62\%). However, this attack still fails to attack our CBD (CA=84.19\%, ASR=4.31\%). More detailed results are shown in Appendix \ref{app:more results}. We can conclude that our defense is resistant to this adaptive attack.
The most probable reason is that the optimized noise becomes less effective when the model is retrained and the model parameters are randomly initialized. In another word, the optimized perturbations are not transferable.

\section{Conclusion}
Inspired by the causal perspective, we proposed Causality-inspired Backdoor Defense (CBD) to learn deconfounded representations for reliable classification. Extensive experiments against 6 state-of-the-art backdoor attacks show the effectiveness and robustness of CBD. Further analysis shows that CBD is robust against potential adaptive attacks. Future works include extending CBD to other domains such graph learning \cite{zhang2022hierarchical,zhang2022protgnn, zhang2021graphmi, zhang2022model}, federated learning \cite{cao2022flcert}, and self-supervised pertaining \cite{zhang2021motif, zhang2021graph}. In summary, our work opens up an interesting research direction to leverage causal inference to analyze and mitigate backdoor attacks in machine learning.

\section*{Acknowledgements}
This research was partially supported by a grant from the National Natural Science Foundation of China (Grant No. 61922073).

{\small
\bibliographystyle{ieee_fullname}
\bibliography{main}
}

\appendix
\clearpage
\newpage
\begin{table*}[ht]
  \caption{Details of datasets and classifiers in the paper}
  \label{table}
  \centering
  \begin{tabular}{ccccc}
    \toprule

      Dataset &  Labels& Input Size& Training Images& Classifier\\
    \midrule
    CIFAR-10 &10 &32 x 32 x 3 &50000 &WideResNet-16-1 \\
    GTSRB& 43& 32 x 32 x 3 &39252& WideResNet-16-1\\
    ImageNet subset &12& 224 x 224 x 3 &12406& ResNet-34\\

    \bottomrule
  \end{tabular}
  \label{datasets}
\end{table*}

\section{More Experimental Settings}
\label{app:settings}
\subsection{Datasets and Classifiers}
The datasets and DNN models used in our experiments are summarized in Table \ref{datasets}.
\subsection{Details of Baseline Implementations}
We implemented the baselines including FP\footnote{https://github.com/kangliucn/Fine-pruning-defense}, MCR\footnote{https://github.com/IBM/model-sanitization}, NAD\footnote{https://github.com/bboylyg/NAD} ABL\footnote{https://github.com/bboylyg/ABL}, and DBD \footnote{https://github.com/SCLBD/DBD} with their open-sourced codes.
For Fine-pruning (FP), we pruned the last convolutional layer of the model. For model connectivity repair (MCR),
we trained the loss curve for 100 epochs using the backdoored model as an endpoint and evaluated the
defense performance of the model on the loss curve. As for the Neural Attention Distillation (NAD), we finetuned the backdoored student network for 10 epochs with 5$\%$ of clean data. The distillation
parameter for CIFAR-10 was set to be identical to the value given in the original paper. We cautiously selected the value of distillation
parameter for GTSRB and ImageNet to achieve the best trade-off between ASR and CA. For ABL, we unlearned the backdoored model using the $\mathcal{L}_{GGA}$ loss with 1$\%$ isolated backdoor examples and a learning rate of 0.0001. For our DBD, we adopt SimCLR as the self-supervised method and MixMatch as the semi-supervised method. The filtering rate is set to $50\%$ as suggested by the original paper.

\subsection{Details of CBD Implementations}
In CBD, $f_C$ is trained on poisoned datasets for 100 epochs using Stochastic Gradient Descent (SGD) with an initial learning rate of 0.1 on CIFAR-10 and the ImageNet subset (0.01 on GTSRB), a weight decay of 0.0001, and a momentum of 0.9. The learning rate is divided by 10 at the 20th and the 70th epoch. $D_\phi$ is set as a MLP with 2 layers. The dimensions of the embedding $r$ and $z$ are set as 64.

\section{More Experimental Results}
\label{app:more results}
\begin{algorithm}[ht]
\caption{Adaptive Attack to CBD}
\textbf{Input}: Model $f_\theta$, poisoned dataset $\mathcal{D}'$, clean dataset $\mathcal{D}$, perturbation range $\epsilon$, number of training iterations $T$, step size $\alpha$, update steps $M$.\\
\textbf{Output}: optimized poisoned dataset $\mathcal{D}'$\\
\vspace{-1em}
\begin{algorithmic}[1]
\STATE Initialize $f_\theta$\\
\FOR{$t= 1, \cdots, T$}
\STATE Draw a mini-batch $\mathcal{B} = \{ (x^{(i)},y^{(i)})\}_{i=1}^n$ from $\mathcal{D}\cup\mathcal{D}'$
\STATE $\theta \leftarrow \theta - \eta \nabla_{\theta}\sum_{(x,y)\in \mathcal{B}} \mathcal{L}(f_\theta(x), y)$\\
\FOR{$(x_i, y_i)$ in $\mathcal{D}'$}
\FOR{$m= 1, \cdots, M$}
\STATE $x_i \leftarrow \Pi_\epsilon(x_i + \alpha\cdot \nabla_x\mathcal{L}(f_{\theta}(x_i), y_i))$\\
\ENDFOR
\ENDFOR
\ENDFOR
\end{algorithmic}\label{algor2}
\end{algorithm}

\subsection{Results of Adaptive Attacks}
\label{adaptive attack appendix}
\begin{figure}[ht]
	\centering
	\subfloat[BadNets]{\includegraphics[width=0.24\linewidth]{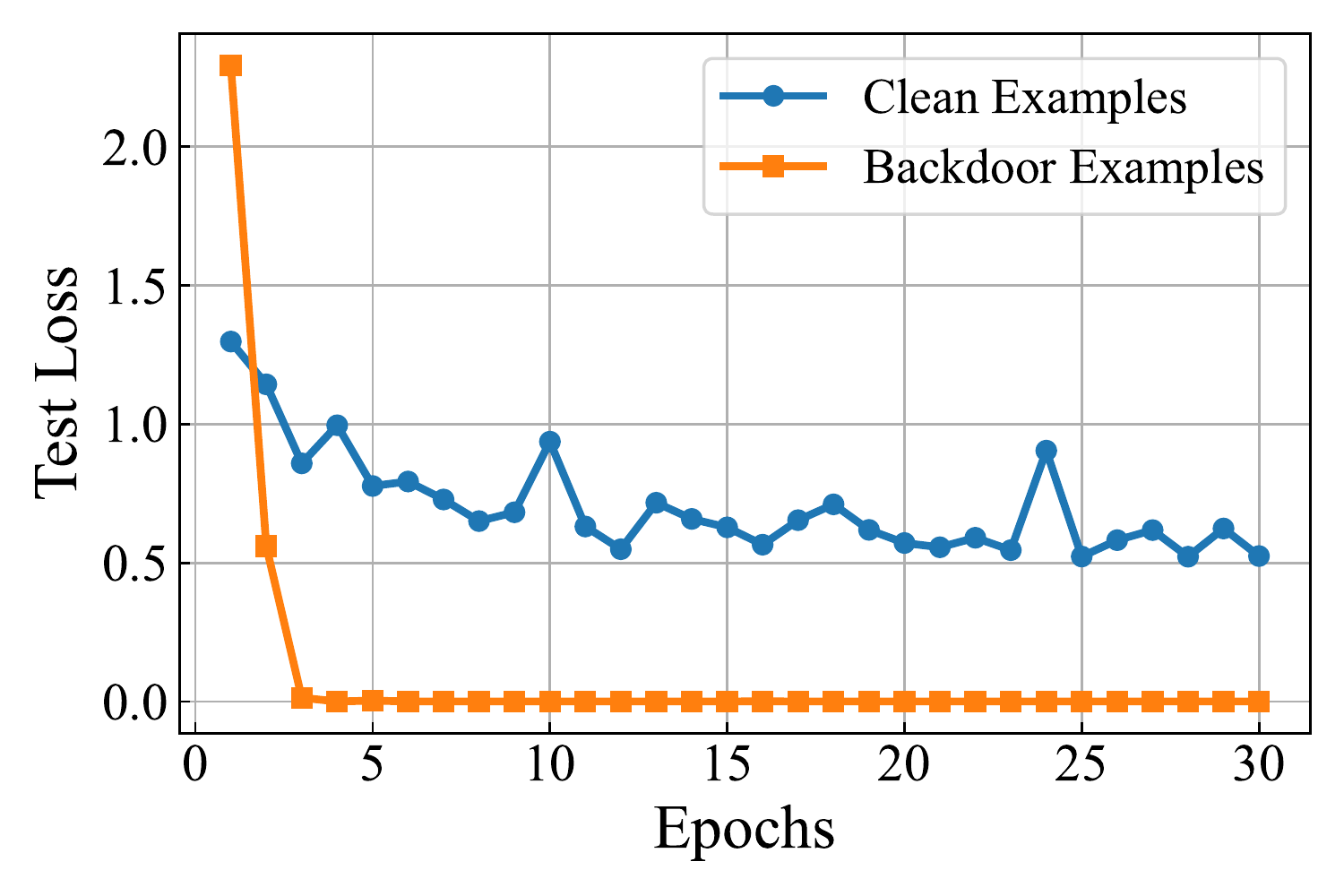}}
    \subfloat[Trojan]{\includegraphics[width=0.24\linewidth]{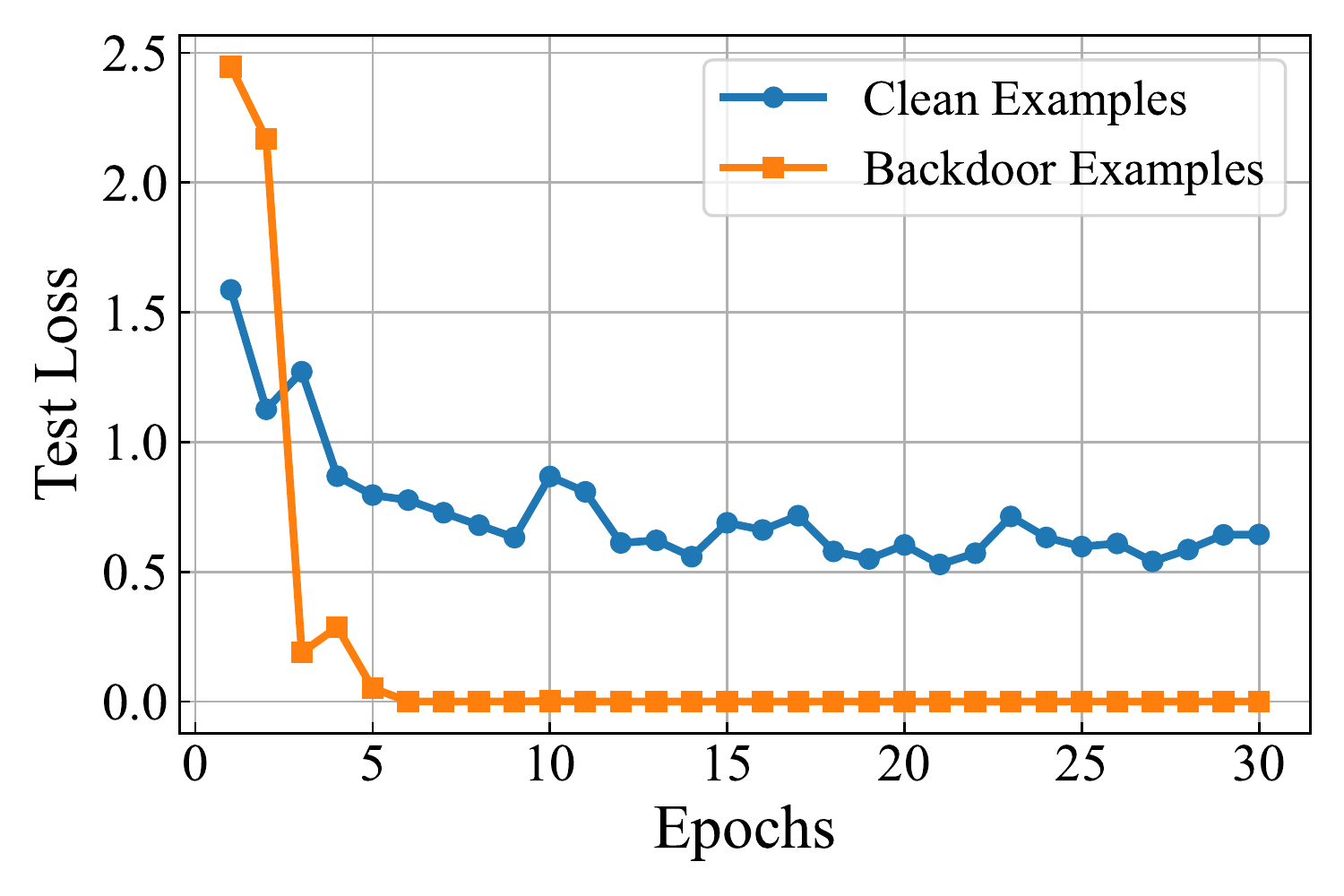}}
    \subfloat[Blend]{\includegraphics[width=0.24\linewidth]{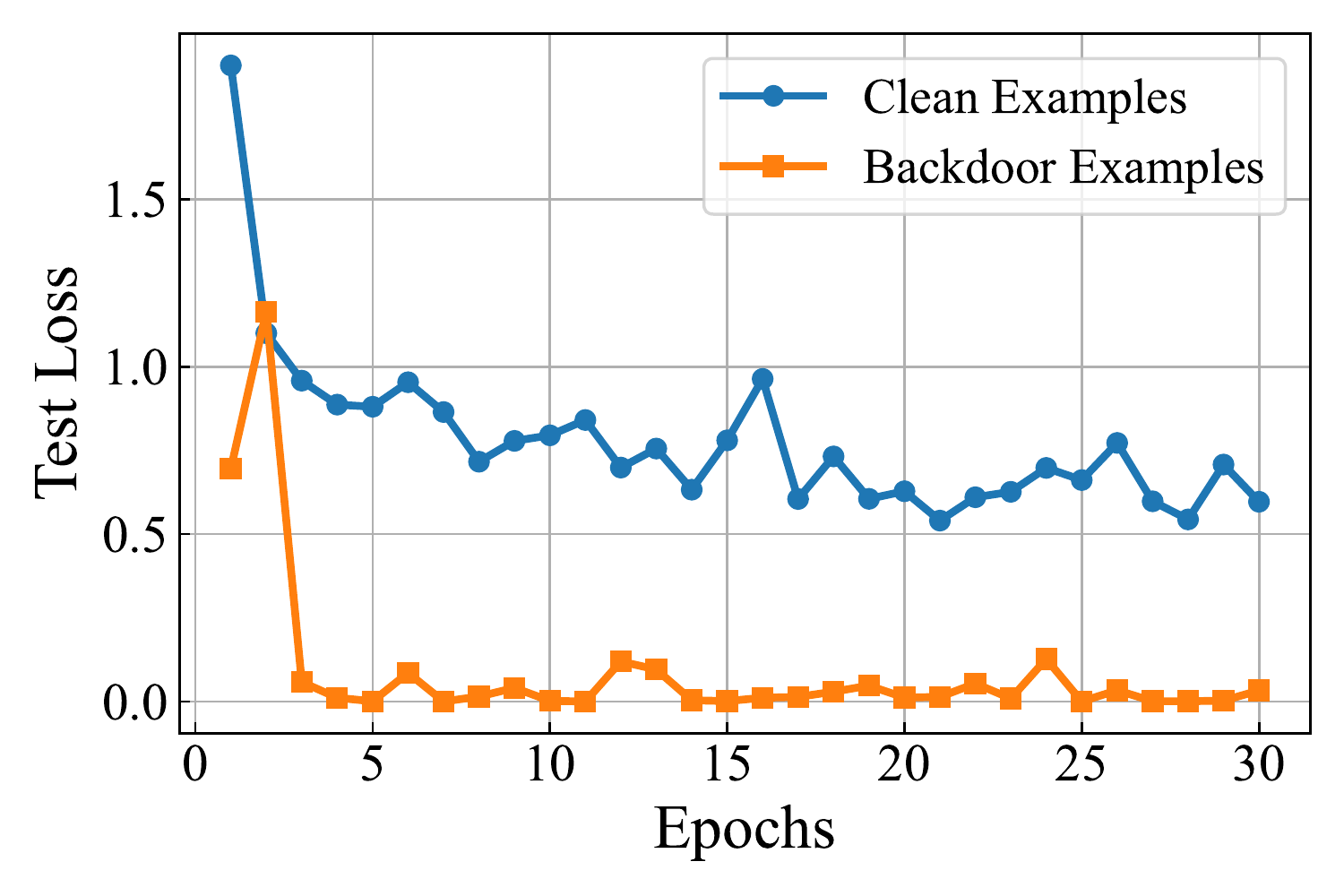}}
    \subfloat[WaNet]{\includegraphics[width=0.24\linewidth]{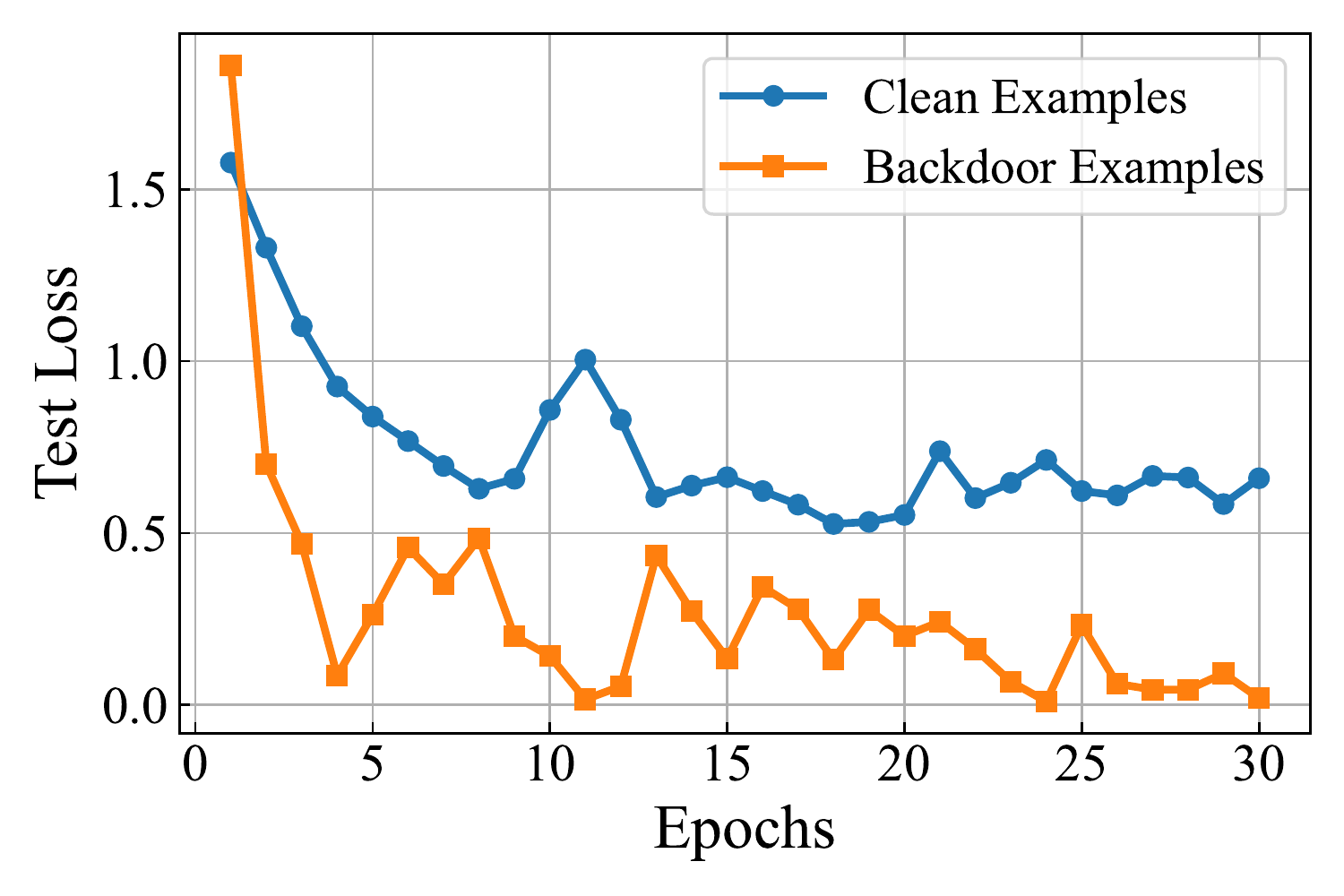}}\\
	\subfloat[BadNets]{\includegraphics[width=0.24\linewidth]{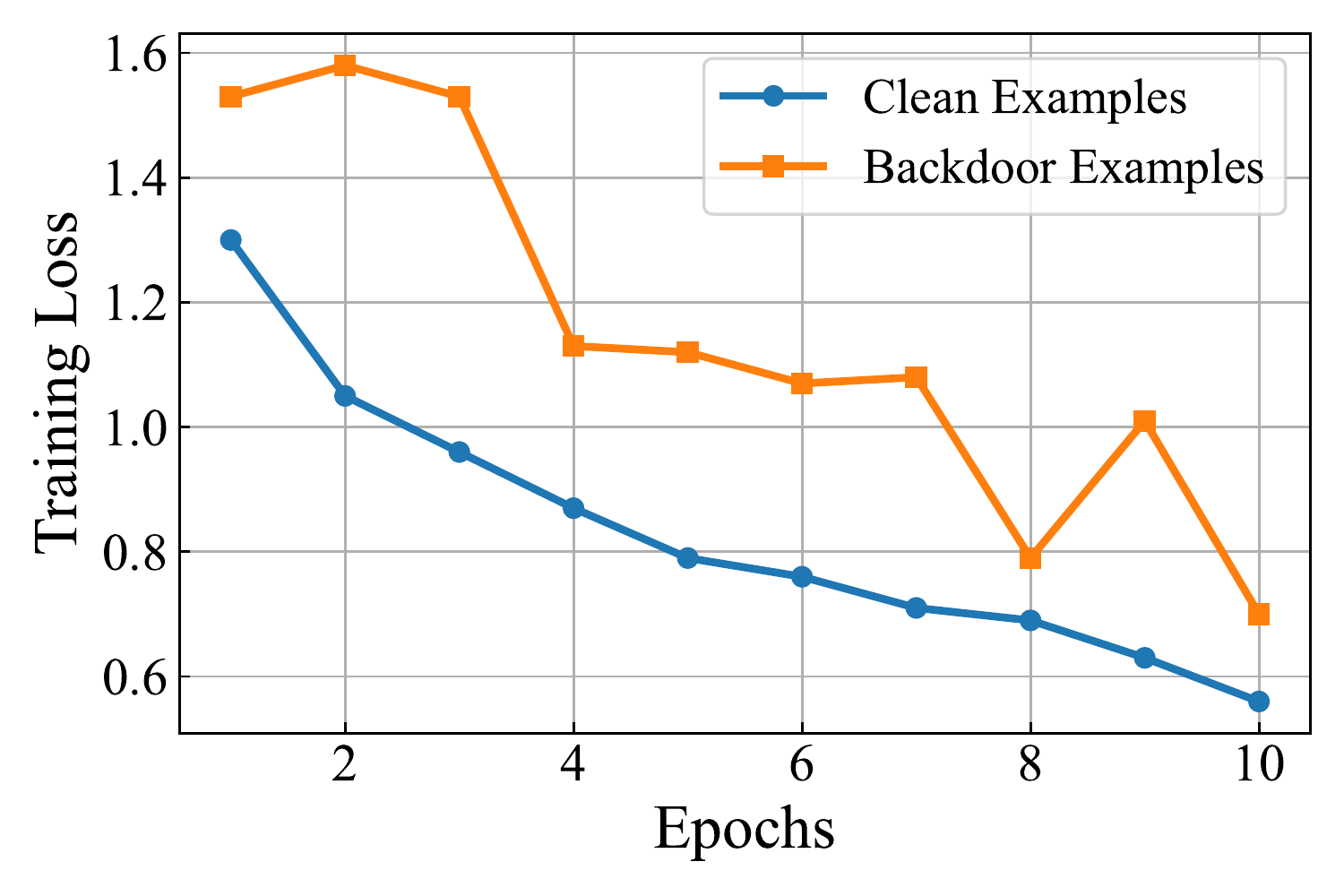}}
    \subfloat[Trojan]{\includegraphics[width=0.24\linewidth]{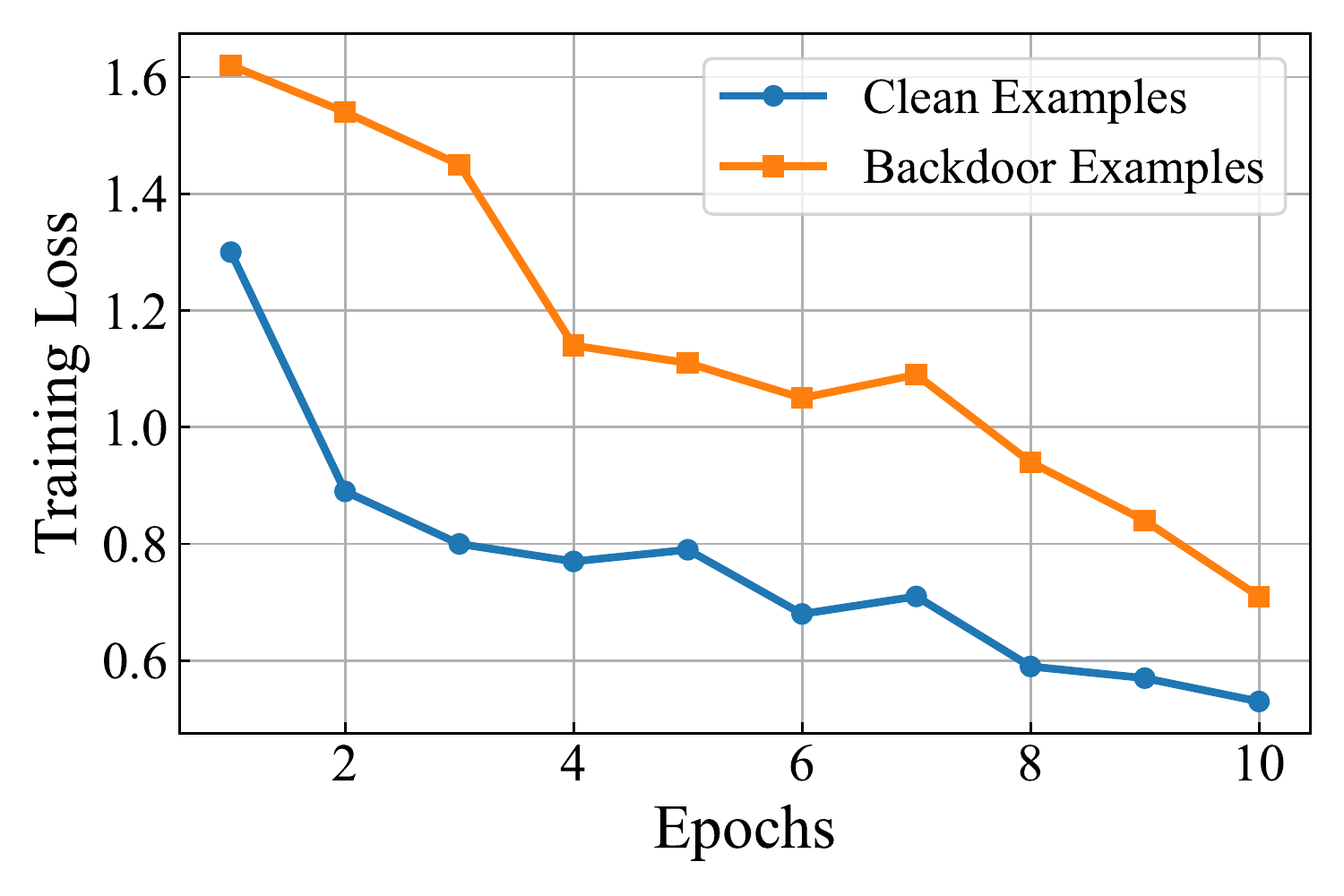}}
    \subfloat[Blend]{\includegraphics[width=0.24\linewidth]{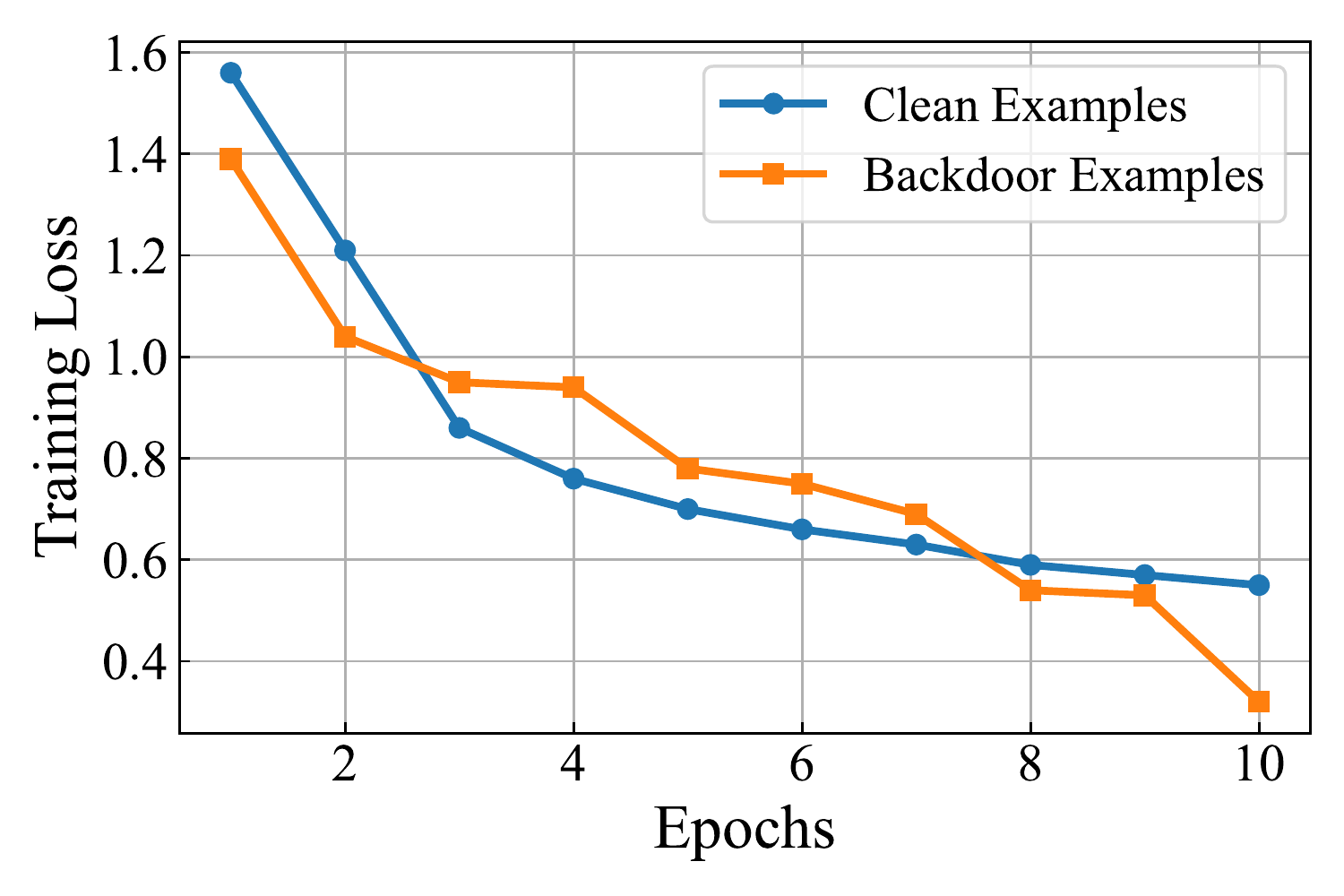}}
    \subfloat[WaNet]{\includegraphics[width=0.24\linewidth]{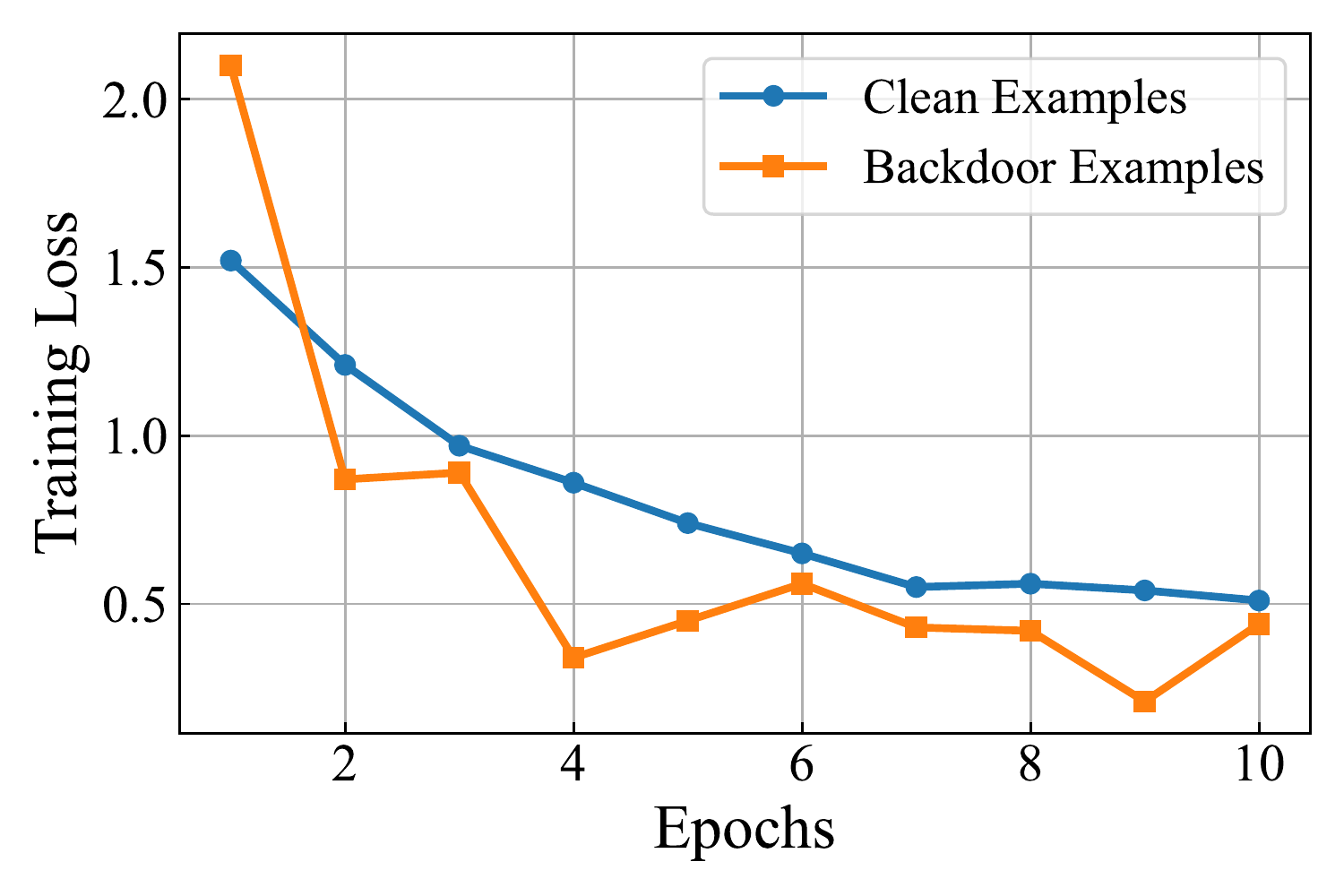}}
	\caption{The curve of training losses on clean/backdoor examples in the vanilla training (first line) and in the optimization of adaptive attacks (second line). This experiment is conducted with WideResNet-16-1 for CIFAR-10 under poisoning rate 10$\%$.}
	\label{adatpive attack curve}
\end{figure}

The pseudo codes of adaptive attacks against CBD are shown in Algorithm \ref{algor2}. The results of adaptive attacks with different kinds of backdoor are shown in Table \ref{adaptive result}. We also show the curves of training losses on clean/backdoor examples in the optimization of added noise along with the vanilla training for reference. In Figure \ref{adatpive attack curve}, we can observe that the training losses of backdoor examples reaches almost zero after several epochs of training (first line) while our adaptive attack strategy managed to increase the losses of backdoor examples in the optimization process (second line). 
The above observation indicates that the backdoor examples are much easier to learn than clean examples in vanilla training. The adaptive attack can slow the injection of backdoor and try to make the backdoor attack stealthier to bypass CBD. 
However, our CBD can still defend the adaptive attack successfully (Table \ref{adaptive result}). The reason is most probably that the optimized noise becomes less effective when the model is retrained and the model parameters are randomly initialized. In another word, the optimized noise is not transferable.

\begin{table}[!tp]
\renewcommand{\arraystretch}{1.2}
\renewcommand\tabcolsep{1.8pt}
\small
\centering
\caption{Attack success rate (ASR $\%$) and clean accuracy (CA $\%$) of Adaptive Attacks.}
\begin{tabular}{c|cccccccc}
\toprule
  \multirow{2}{*}{Defense} & \multicolumn{2}{c}{BadNets} & \multicolumn{2}{c}{Trojan} & \multicolumn{2}{c}{Blend} & \multicolumn{2}{c}{WaNet} \\ \cline{2-9} 
 & ASR & CA & ASR & CA & ASR & CA & ASR & CA \\ \hline
\emph{None} & 99.62 & 84.55 & 99.85 & 84.32 & 97.63 & 84.45 & 97.24 & 85.47 \\  \hline
 CBD & 4.31 & 84.19 & 3.77 & 84.37  & 2.57 & 84.49 & 5.19& 85.33\\ \bottomrule
\end{tabular}
\label{adaptive result}
\end{table}

\subsection{Results with Different Model Architectures}
Note that our CBD is agnostic to the choice of model architectures. In the main text, we report the results with WideResNet-16-1 and ResNet-34. Here, in Table \ref{more architectures} and \ref{vit}, we show experimental results on CIFAR-10 with WideResNet-40-1 \cite{zagoruyko2016wide} and th T2T-ViT \cite{liu2021efficient} under poisoning rate $10\%$. We can observe that CBD can still greatly reduce the attack success rate and keep clean accuracy with different model architectures. 

\begin{table}[ht]
\renewcommand{\arraystretch}{1.2}
\renewcommand\tabcolsep{1.8pt}
\small
\centering
\caption{Results on CIFAR10 with WideResNet-40-1. We show attack success rates (ASR $\%$) and clean accuracy (CA $\%$).}
\begin{tabular}{c|cccccccc}
\hline
  \multirow{2}{*}{Defense} & \multicolumn{2}{c}{BadNets} & \multicolumn{2}{c}{Trojan} & \multicolumn{2}{c}{Blend} & \multicolumn{2}{c}{WaNet} \\ \cline{2-9} 
 & ASR & CA & ASR & CA & ASR & CA & ASR & CA \\ \hline
\emph{None} & 100 & 92.96 & 100 & 93.21 & 99.83 & 92.69 & 98.35 & 92.88\\  \hline
 CBD & 0.95 & 92.54 & 1.04 & 92.70  & 1.32 & 92.17 & 2.54& 92.26\\ \hline
\end{tabular}
\label{more architectures}
\end{table}

\begin{table}[ht]
\renewcommand{\arraystretch}{1.2}
\renewcommand\tabcolsep{1.8pt}
\small
\centering
\caption{Results on CIFAR10 with T2T-ViT. We show the attack success rates (ASR $\%$) and the clean accuracy (CA $\%$).}
\vspace{-1em}
\begin{tabular}{c|cccccccc}
\hline
  \multirow{2}{*}{Defense} & \multicolumn{2}{c}{BadNets} & \multicolumn{2}{c}{Trojan} & \multicolumn{2}{c}{Blend} & \multicolumn{2}{c}{WaNet} \\ \cline{2-9} 
 & ASR & CA & ASR & CA & ASR & CA & ASR & CA \\ \hline
\emph{None} & 100 & 85.65 & 100 & 85.86 & 99.42 & 85.66 & 99.30 & 86.27\\  \hline
 CBD & 0.89 & 86.05 & 0.97 & 86.61  & 1.59 & 85.82 & 3.80 & 85.94\\ \hline
\end{tabular}
\vspace{-1em}
\label{vit}
\end{table}

\subsection{The computational time of other defenses.}
Table. \ref{time cost1} shows the total computational time of defense methods against BadNets. As the methods belong to different categories,
we count the time to train backdoored models for FP, MCR, and NAD for a fair comparison. Generally, the time cost of CBD is acceptable. 
\begin{table}[h]
\renewcommand\tabcolsep{1.8pt}
\caption{The total computational time (seconds) on CIFAR10 with WRN-16-1. The percentages in parentheses indicate the relative increase compared to no defence (\emph{None}).}
\vspace{-1em}
\scriptsize
  \centering
  \begin{tabular}{ccccccc}
    \toprule
     \emph{None} &FP &MCR &NAD& ABL &DBD &CBD (ours) \\ \midrule
     1152& 1515{\tiny(31.5$\%$)}&3445{\tiny(127.4$\%$)} & 1306{\tiny(13.4$\%$)}& 1383{\tiny(20.0$\%$)} & 5280{\tiny(358.3$\%$)} & 1317{\tiny(14.3$\%$)}\\
    \bottomrule
  \end{tabular}
  \label{time cost1}
  \vspace{-2em}
\end{table}

\section{Details of Derivations}
\label{app: derivation}
Here we show the details of derivation with respect to Equ. \ref{kl gaussian}. Since the $p(z|x)= \mathcal{N}(\mu(x),{\rm diag}\{\sigma^2(x)\})$ and $p(x)= \mathcal{N}(0,1)$ are multivariate Gaussian distributions with independent components, we only need to derive the case with univariate Gaussian distributions. For the univariate case, we have:
\begin{equation}
\begin{aligned}
    &D_{\rm KL}(\mathcal{N}(\mu, \sigma^2)||\mathcal{N}(0,1))\\&=\int \frac{1}{\sqrt{2\pi\sigma^2}} e^{-(x-\mu)^2/2\sigma^2}\left({\rm log}\frac{e^{-(x-\mu)^2/2\sigma^2}/\sqrt{2\pi\sigma^2}}{e^{-x^2/2}/\sqrt{2\pi}}\right) dx \\
    &=\int \frac{1}{\sqrt{2\pi\sigma^2}} e^{-(x-\mu)^2/2\sigma^2} {\rm log}\left\{\frac{1}{\sigma}{\rm exp}\left\{\frac{1}{2}[x^2-(x-\mu)^2/\sigma^2]\right\}\right\} dx\\
    &=\frac{1}{2} \int \frac{1}{\sqrt{2\pi\sigma^2}} e^{-(x-\mu)^2/2\sigma^2}[-{\rm log}\sigma^2+x^2-(x-\mu)^2/\sigma^2]dx \\
    &=\frac{1}{2} (-{\rm log}\sigma^2+\mu^2+\sigma^2-1).
    \label{11}
\end{aligned}
\end{equation}
The final equation of Equ. \ref{11} holds because $-{\rm log}\sigma^2$ is a constant; the term $x^2$ is the second order moment of the Gaussian distribution and equals to $\mu^2 + \sigma^2$ after integration; the $(x-\mu)^2$ in the third term calculates the variance and equals to $\sigma^2$ after integration ($-\frac{\sigma^2}{\sigma^2} = -1$). For the results of multivariate Gaussian distributions, we have:
\begin{equation}
\begin{aligned}
    &D_{\rm KL}(p(z|x)||q(z))\\&=D_{\rm KL}(\mathcal{N}(\mu(x),{\rm diag}\{\sigma^2(x)\})||\mathcal{N}(0,1))\\&=\frac{1}{2}\sum_d  (-{\rm log}\sigma_d^2+\mu_d^2+\sigma_d^2-1)\\&=
    \frac{1}{2}||\mu(x)||_2^2+\frac{1}{2}\sum_d(\sigma_d^2- {\rm log}\sigma_d^2 -1).
\end{aligned}
\end{equation}
Therefore, Equ. \ref{kl gaussian} in the main text has been proved.
\end{document}